\documentclass[runningheads]{llncs}

\usepackage{eccvabbrv}

\usepackage{graphicx}
\usepackage{booktabs}
\usepackage{enumitem}
\usepackage{breakcites}
\usepackage[accsupp]{axessibility}  
\usepackage{algorithm}
\usepackage{algorithmic}

\usepackage{hyperref}

\usepackage{sidecap}
\usepackage{soul}

\usepackage{orcidlink}
\usepackage{multirow}
\usepackage{graphicx}
\usepackage{subcaption}
\usepackage{xcolor}
\definecolor{no_adaptation}{HTML}{A8BDD4}
\definecolor{adaptation_wo_mard}{HTML}{D9A0A0}
\definecolor{adaptation_with_mard}{HTML}{EDE0FF}
\usepackage{booktabs}
\usepackage[table]{xcolor}
\usepackage{colortbl}
\usepackage{array}
\usepackage{amsmath}
\definecolor{ourRow}{RGB}{220, 243, 220}        

\newcommand{\venue}[1]{\scriptsize\textcolor{gray!60}{(#1)}}
\usepackage{amssymb} 
\usepackage{wrapfig}

\begin{document}

\title{Real-Time Source-Free Object Detection} 

\titlerunning{Real-Time Source-Free Object Detection}

\author{Sairam VCR\inst{1} \and
Varun Gopal\inst{1} \and
Poornima Jain\inst{1} \and \\
Vineeth N Balasubramanian\inst{1,2} \and
Muhammad Haris Khan\inst{3}}

\authorrunning{Sairam VCR et al.}

\institute{$^1$IIT Hyderabad, India \quad $^2$Microsoft Research \quad $^3$MBZUAI \\
\email{\{ai20resch13001@,co22btech11015@,ai24resch11002@, \\ vineethnb@cse.\}iith.ac.in}, \email{muhammad.haris@mbzuai.ac.ae}}

\maketitle

\begin{abstract}
Real-world detectors for autonomous driving, surveillance, and robotics must handle domain-shifts under strict latency and memory constraints, yet existing source-free object detection (SFOD) methods rely on heavyweight architectures that prioritize accuracy alone. We show this trade-off is unnecessary: building on YOLOv10, an NMS-free dual-head detector, we achieve state-of-the-art adaptation accuracy while being faster and more compact. We observe that directly applying vanilla mean-teacher self-training to dual-head detectors leads to suboptimal adaptation performance due to two key factors. First, simple pseudo-label generation strategies, such as using a single head or directly combining high-confidence predictions from both heads, yield suboptimal supervision under domain-shift. We propose DHF (Dual-Head Pseudo-Label Fusion) which selectively admits one-to-one (O2O) and one-to-many (O2M) head predictions, preserving precision and recovering missed objects. Second, we observe domain-shift collapses multi-scale feature discriminability. We propose the use of our MARD (Multi-scale Adaptive Representation Diversification) loss which mitigates this by enforcing detection-aware variance and covariance constraints on multi-scale feature maps. Both modules are training-time only, leaving inference unchanged. Across domain-shift benchmarks, our method, RT-SFOD yields 1.4 to 3.5\% mAP gains, 1.3$\times$ higher throughput, with $\sim$2$\times$ fewer parameters than prior state-of-the-art SFOD methods, thus advancing the Pareto frontier of the speed-accuracy-model size trade-off. We report main results with YOLOv10, and demonstrate generalizability with additional YOLO- and DETR-based dual-head detectors. Code is available here: \href{https://github.com/Sairam13001/RT-SFOD/}{https://github.com/Sairam13001/RT-SFOD/}.
\end{abstract}

\section{Introduction}
\label{sec:intro}
\begin{figure}[t]
    \centering
    \includegraphics[width=0.75\linewidth]{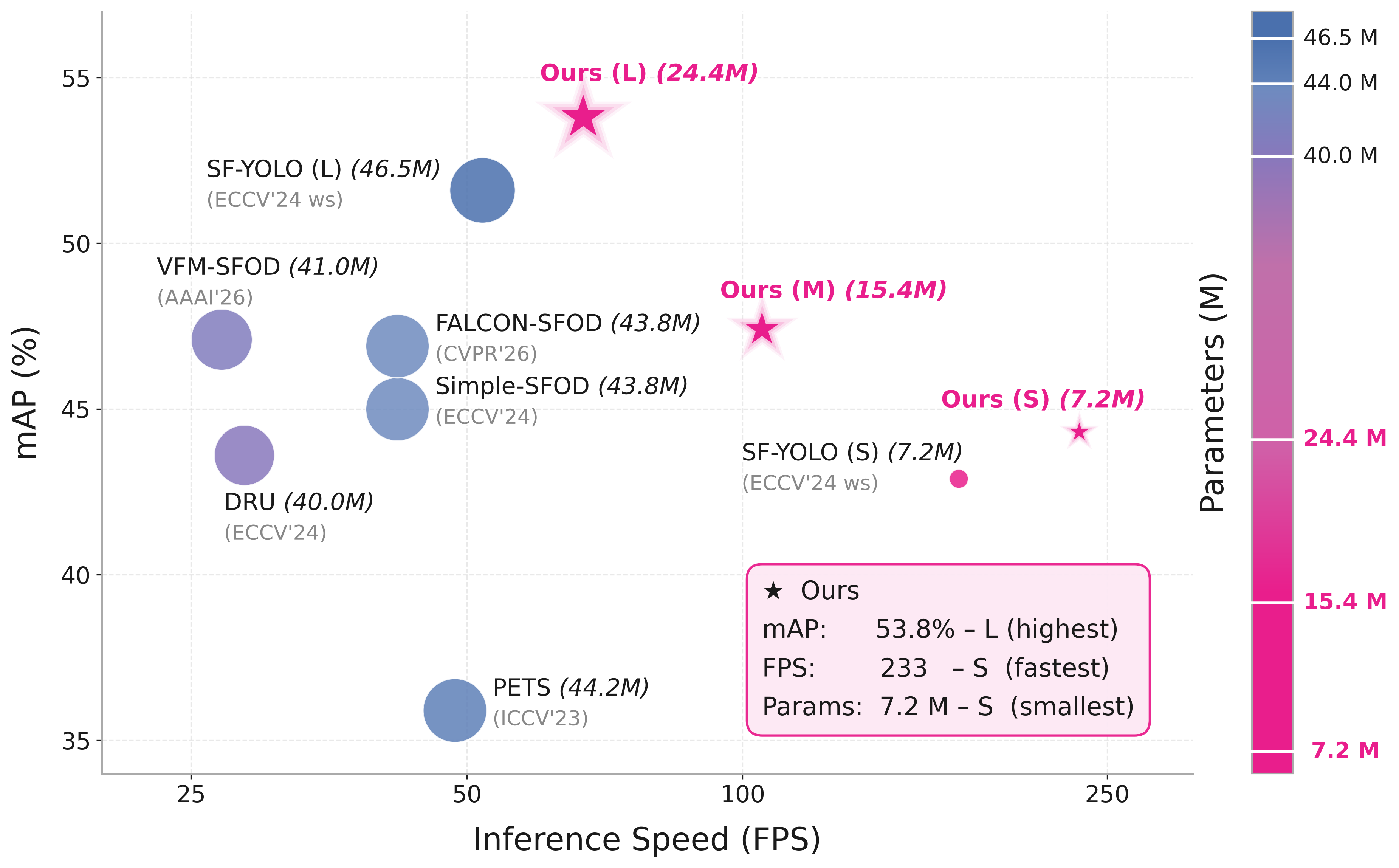}
    \vspace{-10pt}
    \caption{\textbf{Accuracy-speed-size trade-off in SFOD.} We compare state-of-the-art SFOD methods on Cityscapes $\to$ Foggy Cityscapes domain-shift, in terms of mAP (\%) and inference speed (FPS), with model size (\# parameters) encoded by color-graded bubble/star size. Our method achieves the best trade-off across all model-scales, obtaining 53.8\% mAP with the large model while remaining the fastest (233 FPS) and most compact (7.2 M params) with the small model.}
\label{fig:teaser}
\vspace{-16pt}
\end{figure}

\begin{figure}[t]
  \centering
  \subfloat[Pseudo-label quality]{\includegraphics[width=0.48\linewidth]{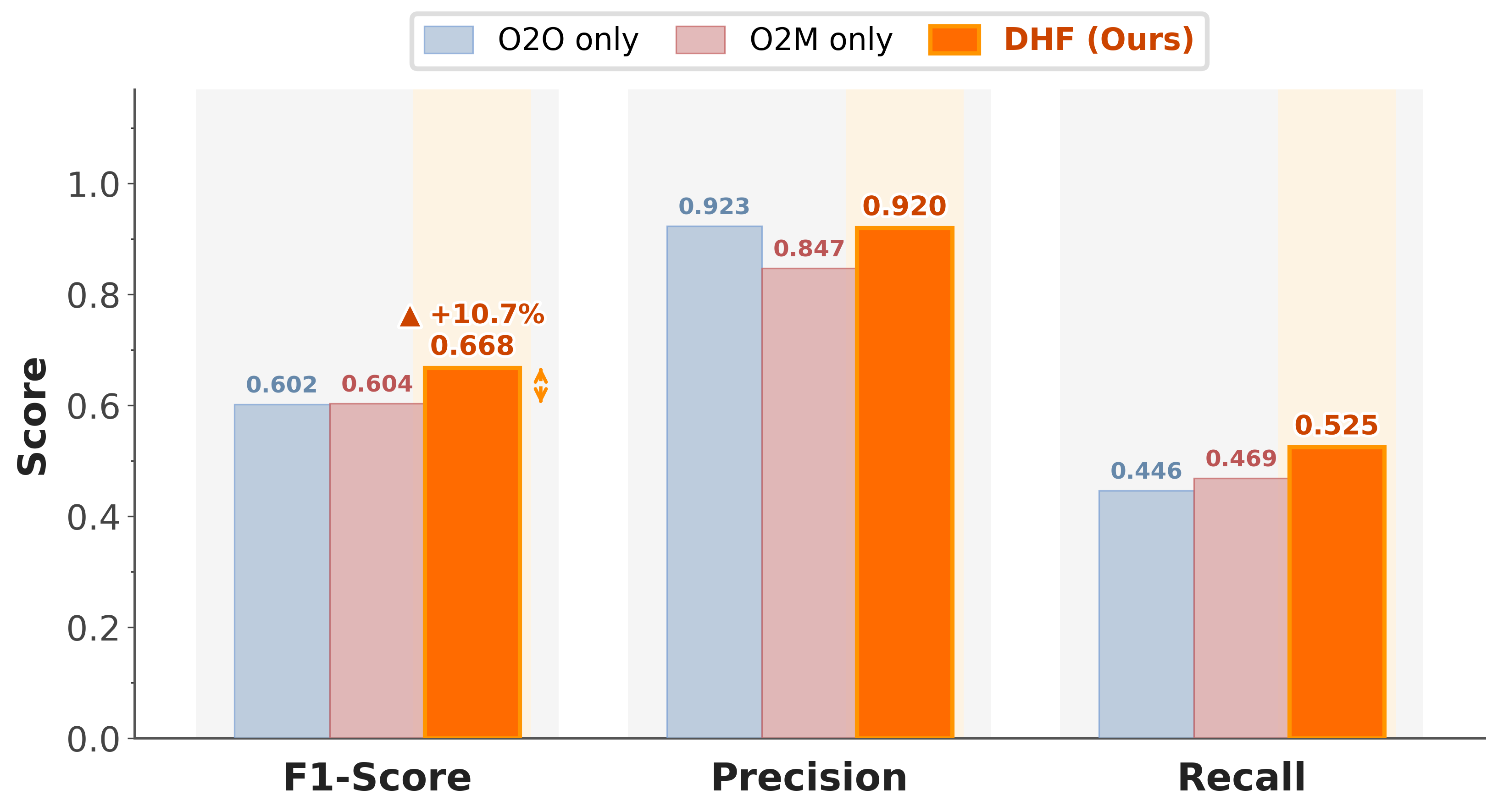}\label{fig:pl_analysis}}
  \hfill
  \subfloat[Feature discriminability]{\includegraphics[width=0.48\linewidth]{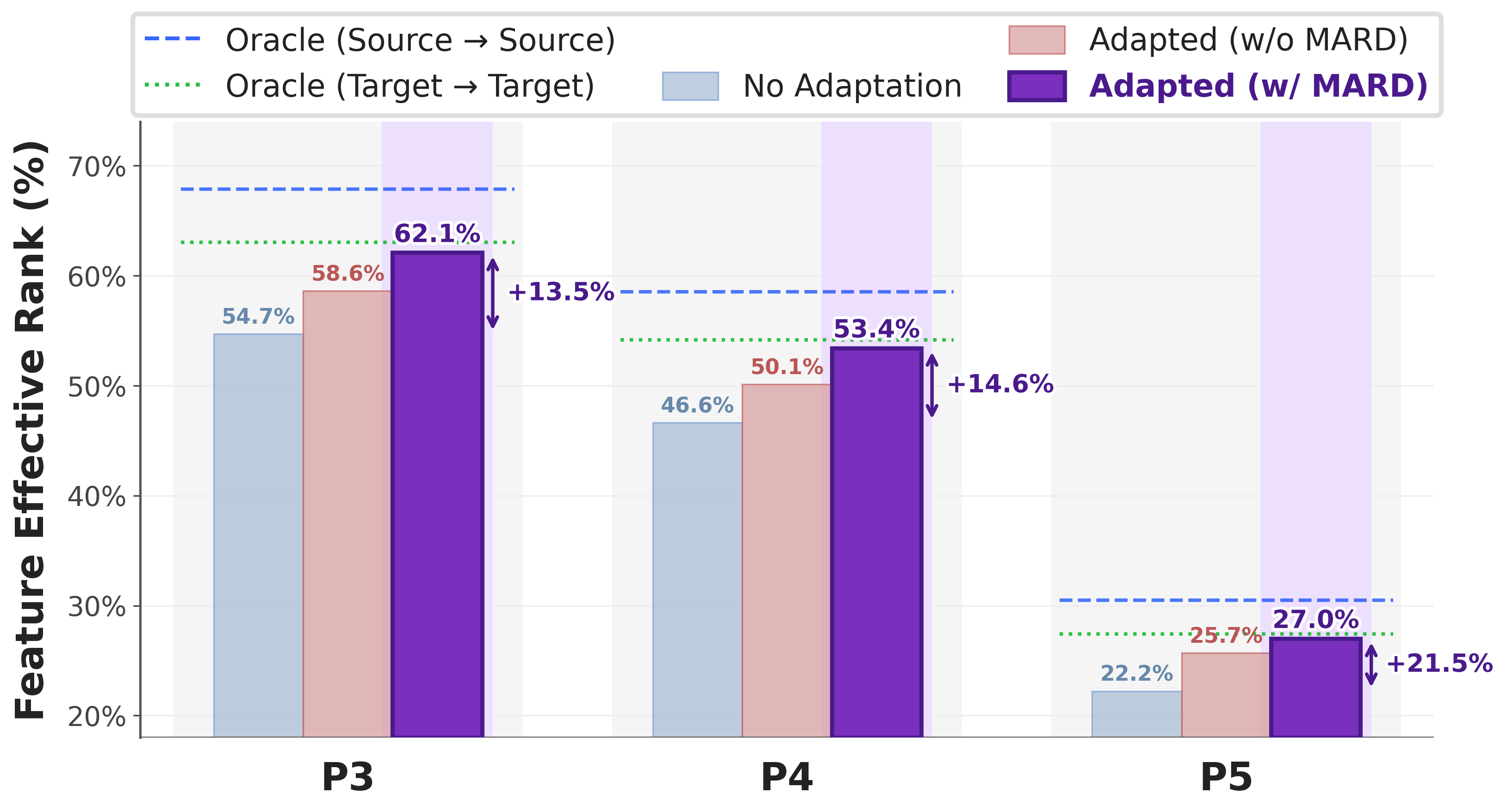} \label{fig:feat_analysis}}
  \caption{\textbf{Motivation for our proposed modules} on the Cityscapes $\to$ Foggy Cityscapes benchmark. \textbf{(a)}~O2O pseudo-labels are precise but miss many objects; O2M pseudo-labels with standard post-processing provide broader coverage but introduce additional noise; our proposed DHF achieves the best F1-score. \textbf{(b)} Effective rank of YOLOv10 multi-scale PAN features (P3, P4, P5) under domain-shift. The dashed blue line marks the source oracle: effective rank of the source model on Cityscapes val set feature maps. The dotted green line marks the target oracle: effective rank of the target model on Foggy Cityscapes val set feature maps. \colorbox{no_adaptation}{\scriptsize{(No Adaptation)}} denotes the effective rank of the source model on Foggy Cityscapes val set feature maps. MARD consistently recovers the effective rank best across all scales, closing the gap toward the target oracle.}
  \label{fig:motivation}
  \vspace{-24pt}
\end{figure}

Source-Free Object Detection (SFOD) adapts a detector trained on labeled source data to an unlabeled target domain \emph{without} using source samples \cite{irg, simple-sfod, falcon-sfod, beyond_boundaries}. This setting is increasingly relevant in applications such as autonomous driving, surveillance, robotics, and medical imaging, where privacy concerns, proprietary restrictions, or storage limitations preclude access to source data\cite{sun2022shift, safdari2025mixstyleflow, noori2026histopath, aljalbout2025reality}. Such real-time applications typically operate under strict latency and memory constraints\cite{ma2024slowperception, sinha2024towards}. However, following the pioneering SFOD method \cite{sfod1_sed}, existing works\cite{irg, pets, simple-sfod, franck, falcon-sfod} largely build on two-stage Faster R-CNN \cite{faster-rcnn}, with a few recent ones\cite{dru, franck, titan, beyond_boundaries} relying on transformer-based detectors \cite{deformable-detr, focal_modulation_networks}. While these backbones are strong, they are ill-suited for real-time applications \cite{rtdetrv3, rf-detr}. One-stage detectors such as YOLO \cite{yolov5, yolov8, yolov10} are designed for efficiency, but limited attention has been given to developing SFOD methods based on them \cite{sf-yolo}. To bridge this gap, we build a new SFOD framework on YOLOv10\cite{yolov10}, which beats the existing methods on accuracy while using a fraction of their parameter count, thus establishing a new state-of-the-art on the accuracy-speed-size frontier as shown in Fig. \ref{fig:teaser}. While we adopt YOLOv10 as the primary architecture for all main experiments, our framework applies to any NMS-free dual-head detector, as we validate on other YOLO- and DETR-based detectors in Table \ref{tab:generality}.

YOLOv10 is the first non-maximum suppression (NMS)-free YOLO detector, employing a dual-head training strategy with one-to-many (O2M) and one-to-one (O2O) heads under a consistent matching scheme to enable true end-to-end, real-time detection without post-processing overhead, while maintaining high accuracy. This makes YOLOv10 appear to be an ideal backbone for real-time SFOD; however, unlike Faster R-CNN and transformer-based detectors, it does not explicitly generate region proposals or object queries that provide structured object-level representations for cross-domain alignment, thereby making source-free adaptation more challenging.


The Mean-Teacher (MT) framework \cite{mean_teacher} facilitates training without ground-truth labels through teacher-student consistency, and has therefore become a fundamental component of state-of-the-art SFOD methods \cite{simple-sfod, franck, falcon-sfod, beyond_boundaries}. However, we observe that a vanilla MT on YOLOv10 is suboptimal for two reasons: (i) As shown in Fig. \ref{fig:pl_analysis}, the O2O head produces high-precision (0.923) pseudo-labels but tends toward low recall (0.446), missing objects. In contrast, use of the O2M head for pseudo-labeling improves coverage (recall 0.469) but introduces additional noise (precision 0.847). Using either head alone for pseudo-labeling, or directly combining head predictions remains suboptimal (Tab. \ref{tab:dhf_ablation}). 
(ii) The dual-head consistent assignment mechanism benefits from highly discriminative features \cite{yolov10}, however, we hypothesize that domain-shift erodes the discriminability. To verify this, for each scale level $\ell$, we reshape the feature map $F_\ell \in \mathbb{R}^{C_\ell \times H_\ell \times W_\ell}$ into a spatially flattened matrix $Z_\ell \in \mathbb{R}^{(H_\ell W_\ell) \times C_\ell}$, where rows correspond to spatial locations and columns correspond to channels. We compute singular values $\{\sigma_i\}$ of $Z_\ell$, normalize them as $p_i = \sigma_i / \sum_j \sigma_j$, and define effective rank as $\mathrm{erank}_\ell = \exp\!\left(-\sum_i p_i \log p_i\right)$, which quantifies the effective number of channel dimensions carrying independent information. Over a dataset of $N$ images, we calculate the dataset-level mean effective rank as $\overline{\mathrm{erank}}_\ell = \frac{1}{N}\sum_{n=1}^{N} \mathrm{erank}^{(n)}_\ell$, and report its normalized percentage as $\mathrm{erank}\%_\ell = 100 \times \frac{\overline{\mathrm{erank}}_\ell}{C_\ell}$ in Fig.\ref{fig:feat_analysis}. Domain-shift erodes the effective rank of features across scales as observed in Fig.\ref{fig:feat_analysis} \colorbox{no_adaptation}{\scriptsize{(No Adaptation)}}, indicating the collapse of features toward lower-dimensional subspaces. Standard mean-teacher self-training \colorbox{adaptation_wo_mard}{\scriptsize{(Adapted w/o MARD)}} recovers less than half of this lost diversity, placing a ceiling on adaptation performance.

To address these two bottlenecks, we introduce two modules respectively: \textbf{(DHF) Dual-Head Pseudo-Label Fusion.} Treating duplicate-free high-precision O2O predictions as anchors, we \emph{selectively} supplement them with non-redundant, high-scoring O2M predictions to recover objects that O2O misses. Fig.\ref{fig:pl_analysis} shows how our \emph{DHF loss increases F-1 score by 10.7\%}, leading to a gain in adaptation performance (Tab.\ref{tab:dhf_ablation}).  This fusion is applied \emph{only} during training-time pseudo-label construction; at inference, the detector retains its original NMS-free O2O-only inference with no additional overhead. 
\textbf{(MARD) Multi-scale Adaptive Representation Diversification.}
MARD enforces structured representational constraints on the multi-scale features, preserving the discriminative capacity (Fig.\ref{fig:feat_analysis} \colorbox{adaptation_with_mard}{\scriptsize{(Adapted w/ MARD)}}) that domain-shift otherwise erodes.
Together, these components facilitate stable adaptation of YOLOv10, achieving similar or higher accuracy than state-of-the-art approaches, with 1.3x higher FPS and requiring significantly fewer parameters. Our main contributions are:


\begin{itemize}[topsep=0pt]
\item We present, to the best of our knowledge, the first systematic study of NMS-free dual-head detectors for SFOD. Our proposed method, \emph{RT-SFOD}, establishes a new state-of-the-art on the accuracy-speed-size frontier.
\item We propose DHF, a training-time fusion of high-precision O2O and selectively filtered O2M predictions for superior pseudo-label quality, and MARD, a multi-scale feature diversification mechanism that counteracts feature rank degradation under domain-shift.
\item Across domain-shift scenarios, our proposed modules yield state-of-the-art or competitive adaptation accuracy, while the efficient YOLOv10 backbone contributes 1.3x higher throughput with $\sim$2x fewer parameters than prior SFOD methods.
\end{itemize}

\section{Related Works}
\label{sec:related_works}
\vspace{-6pt}
\noindent\textbf{Source-Free Object Detection: }SFOD was introduced in \cite{sfod1_sed}, which identified noisy pseudo-labels as the main challenge and addressed it using entropy-based confidence thresholding and false-negative augmentation. LODS \cite{LODS} applied bidirectional knowledge distillation on original and style-transferred target images with graph-based feature alignment to learn domain-invariant representations. \cite{Chu_Li_Chen_Li_Li_2023} proposed A$^2$SFOD, which uses variance-based target splitting and adversarial alignment in a mean-teacher framework. IRG~\cite{irg} adopted a mean-teacher framework with graph CNN-guided contrastive learning for target feature alignment. Simple-SFOD \cite{simple-sfod} adapted batch normalization statistics and leveraged strong augmentations with fixed pseudo-labels in a teacher-student setup for domain adaptation. FALCON-SFOD \cite{falcon-sfod} showed that domain-shift weakens object-focus in features and leveraged foundation model priors to address this issue. Following \cite{sfod1_sed}, all above methods adopt Faster R-CNN as the base detector. Recent SFOD \cite{dru, franck} works utilize Detection Transformers (DETRs). In \cite{dru}, a historical student loss is used to stabilize student-training and mitigate noisy pseudo-labels. FRANCK \cite{franck} is an SFOD framework that enhances DETR adaptation via category-level contrastive learning, instance reweighting, and uncertainty-weighted feature distillation. 
Faster R-CNN and DETR are computationally intensive~\cite{rcnn_detr_yolo_efficiency}, and SFOD methods built on them focus mainly on adaptation accuracy, overlooking real-time performance metrics such as FPS and model size. 

\noindent\textbf{Unsupervised Domain Adaptive Object Detection:}
UDAOD addresses domain-shift using both labeled source and unlabeled target data. Early methods focused on Faster R-CNN~\cite{fr1,fr2,fr3,fr4}, while later works adopted DETR-based models: MTM~\cite{mtm} proposed two-stage DETR adaptation with source pre-training on style-transferred images and masked feature alignment; MTTrans~\cite{mttrans} applied a ResNet50-based DETR in a mean-teacher framework; DA-DETR~\cite{dadetr} and SFA~\cite{sfa} introduced transformer detectors with dedicated alignment modules; and BiADT~\cite{liu2022dabdetr} leveraged box-coordinate queries as positional priors using a DAB-DETR backbone.
UDOAD requires access to labeled source data, which limits its applicability in real-world applications where source data privacy and unavailability are major concerns, shifting the focus more towards SFOD.

\noindent\textbf{Efficiency-centric Object Detection:}
Research on fast/lightweight detectors includes SSD-MobileNet~\cite{liu2016ssd, howard2017mobilenets} and EfficientDet~\cite{efficientdet}. NanoDet~\cite{nanodet} adopts depthwise convolutions with an anchor-free design, while the YOLO line~\cite{yolov5, yolov7, yolov10, yolov11} has been progressively optimized for speed. Real-time transformer detectors include Real-Time-DETR~\cite{rtdetr}, RF-DETR~\cite{rf-detr}, LightWeight-DETR~\cite{lw-detr}, and Quantized RT-DETR~\cite{qrt-detr}, but they generally trail YOLO in throughput. For instance,~\cite{yolov10} reports RT-DETR-R18 at $\sim$46.5 AP on COCO with $\sim$4.6 ms/frame ($\sim$217 FPS), whereas YOLOv10-S achieves similar AP (46.3) in 2.49 ms/frame ($\sim$401 FPS), about \emph{1.8$\times$ faster}. The same study~\cite{yolov10} also shows Faster R-CNN and DETR variants have higher compute/latency, while YOLOv10’s streamlined heads and lightweight blocks surpass YOLOv8 at higher accuracy with $\sim$2$\times$ fewer parameters, making it especially efficient for real-time systems.

SF-YOLO~\cite{sf-yolo} is the first work to apply real-time detectors to SFOD. It adapts a single-head, NMS-based YOLO detector (YOLOv5~\cite{yolov5}) using mean-teacher self-training with student stabilization and learned augmentations. In contrast, our work focuses on NMS-free dual-head detectors, which bring new challenges beyond those considered in SF-YOLO, such as pseudo-label precision-recall trade-off across heads. We further provide the evidence of a previously unreported phenomenon: feature discriminability collapse under domain-shift; and provide a scale-aware regularization loss to mitigate this effect and preserve feature diversity. To our knowledge, no prior work systematically studies efficiency-driven SFOD for NMS-free detectors, leaving newer efficient architectures such as YOLOv10~\cite{yolov10} underexplored. We fill this gap with an SFOD framework for NMS-free dual-head detectors, instantiated primarily on YOLOv10, where our proposed modules drive adaptation accuracy gains while the throughput and latency advantages stem from the efficient YOLO backbone. We also demonstrate generality on other YOLO- and DETR-based dual-head detectors.

\vspace{-8pt}
\section{Methodology}
\label{sec:method}
\begin{figure}[t]
    \centering
    \includegraphics[width=\linewidth]{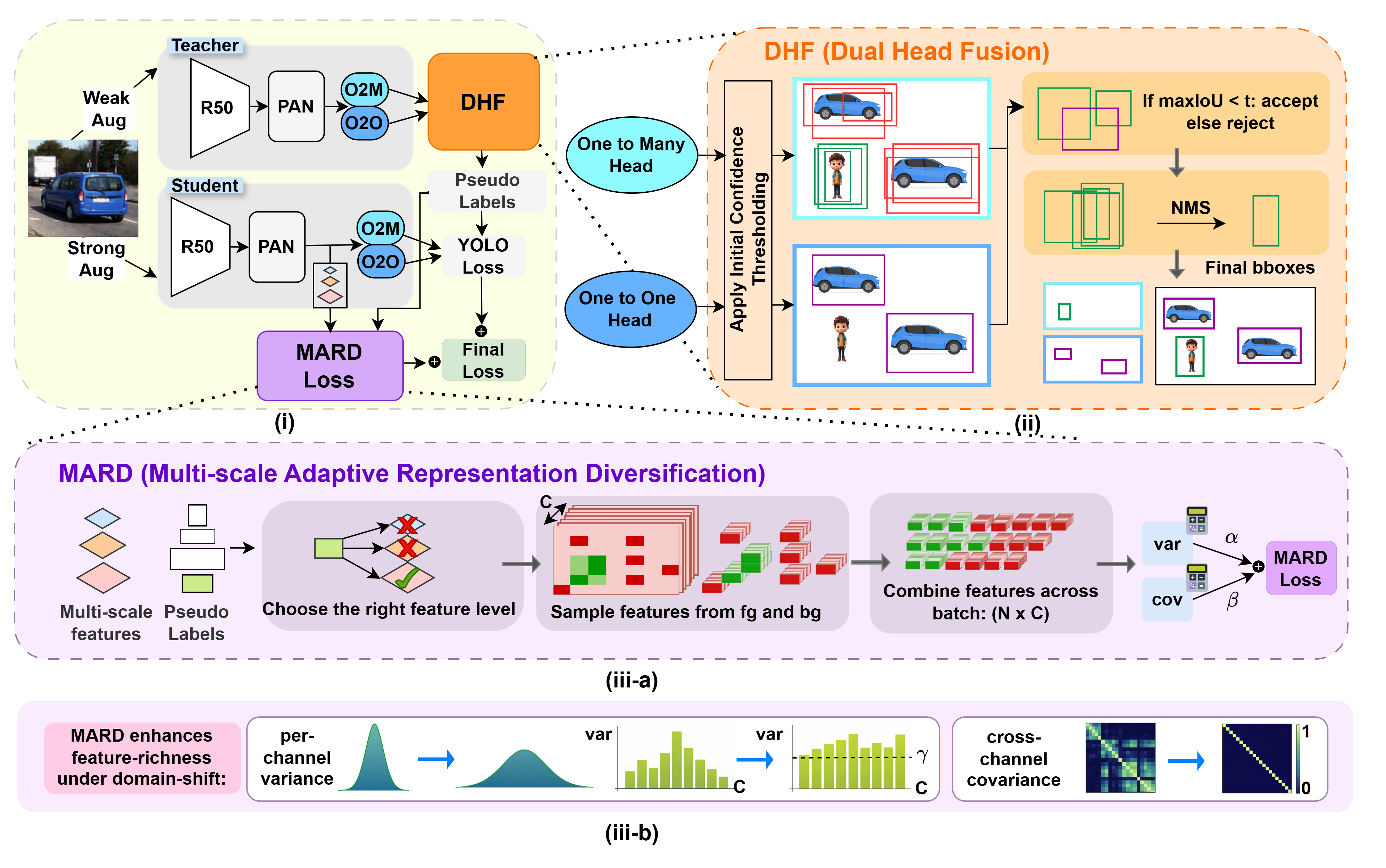}
    \vspace{-12pt}
    \caption{\textbf{Overview of RT-SFOD.} (i) Mean-Teacher framework with DHF and MARD.
(ii) \textbf{DHF:} High-precision O2O predictions serve as anchors; non-redundant O2M boxes
with low overlap w.r.t.\ O2O anchors are selectively added to form the final pseudo-labels.
(iii) \textbf{MARD:} (iii-a) Foreground and background feature vectors are sampled from multi-scale features using pseudo boxes and regularized via variance and covariance constraints,
(iii-b) increasing per-channel variance for better discriminability and reducing off-diagonal
covariance for greater channel diversity.}
\label{fig:architrecture}
\vspace{-12pt}
\end{figure}

\subsection{Preliminaries}
\noindent\textbf{Problem statement.} Let the unlabeled target dataset be $\mathcal{D}_t=\{x_i\}_{i=1}^{N}$. 
Given a detector pre-trained on a labeled source domain, our objective is to adapt it to $\mathcal{D}_t$ \emph{without} accessing any source samples, while maintaining a favorable accuracy-efficiency trade-off suitable for real-time use.

\noindent\textbf{Mean-Teacher self-training.} Most state-of-the-art SFOD methods follow the Mean-Teacher (MT) teacher-student paradigm. A teacher detector $f_{\theta_T}$ produces pseudo-labels on the target domain, which are then used to supervise a student detector $f_{\theta_S}$. 
For a target image $x$, we construct a weakly augmented view $x^w=a_w(x)$ and a strongly augmented view $x^s=a_s(x)$. The teacher predicts pseudo-labels on $x^w$, and the student is trained to match them on $x^s$. 
After each step, the teacher parameters are updated via an exponential moving average (EMA) of the student with momentum $\mu$: $\theta_T \leftarrow \mu\,\theta_T + (1-\mu)\,\theta_S.$

\noindent\textbf{YOLOv10 preliminaries.} We adopt YOLOv10 as the primary base detector due to its strong accuracy-efficiency trade-off and end-to-end NMS-free inference, which are desirable for real-time SFOD. YOLOv10 employs two prediction heads trained under a consistent matching scheme: a \emph{one-to-one} (O2O) head that assigns each object to a single prediction, and a \emph{one-to-many} (O2M) head that allows multiple matched predictions per object to strengthen supervision. YOLOv10 operates on a multi-scale feature pyramid (PAN); we denote the student features at pyramid level $l$ as $F_\ell^S$, for $\ell\in\{P3,P4,P5\}$. During inference, only the O2O head is used to produce duplicate-free detections.

\noindent\textbf{Why vanilla MT is suboptimal for YOLOv10.} Despite being a strong SFOD baseline, vanilla MT underperforms on YOLOv10 for two reasons. \textbf{(i)} Its dual heads yield complementary but mismatched pseudo-labels: O2O is duplicate-free and precise yet misses many objects under domain-shift, whereas use of the O2M head for pseudo-labeling improves coverage but introduces additional noise; using either alone, or directly combining them, results in suboptimal supervision. \textbf{(ii)} Domain-shift degrades YOLOv10’s multi-scale feature discriminability\cite{yolov10}, reducing effective rank across pyramid levels (Fig.~\ref{fig:feat_analysis}), and MT self-training only partially recovers this diversity. We therefore propose DHF for higher-quality pseudo-labels and MARD to preserve discriminative features.

\vspace{-8pt}
\subsection{Proposed Real-Time SFOD}
\label{sec:mt} 
\vspace{-4pt}
Following \cite{simple-sfod}, we warm-start the teacher using Adaptive Batch Normalization (AdaBN): we update BN statistics on target images to obtain a stronger target-initialized checkpoint. Teacher and student are initialized with this checkpoint before the adaptation begins. After applying AdaBN, we found that updating the teacher model once per epoch yields better performance than updating it at every training step. \noindent Algorithm~\ref{alg:overall} summarizes the adaptation loop. Below, we explain the two proposed modules of our framework. 

\begin{algorithm}[t]
\caption{Proposed Real-time SFOD.}
\label{alg:overall}
\begin{algorithmic}[1]
\REQUIRE AdaBN-initialized checkpoint $\theta^0$, unlabeled target data $\mathcal{D}_t$ and hyperparameters
\STATE Initialize student $\theta_S\leftarrow\theta^0$, teacher $\theta_T\leftarrow\theta^0$
\FOR{each epoch}
  \FOR{each minibatch $x\sim\mathcal{D}_t$}
    \STATE Form $x^w=a_w(x)$ and $x^s=a_s(x)$
    \STATE Teacher predicts $\mathcal{P}^{o}$ and $\mathcal{P}^{m}$ on $x^w$
    \STATE Fuse with DHF to obtain pseudo-labels $\hat{Y}^w$ \hfill (Eq. \ref{eq:DHF})
    \STATE Map $\hat{Y}^w \rightarrow \hat{Y}^s$ via the strong-view geometry
    \STATE Student predicts $Y$ on $x^s$; compute detection loss $\mathcal{L}_{\text{det}}(\hat{Y}^s, Y)$
    \STATE Compute MARD loss $\mathcal{L}_{\text{mard}}$ on multi-scale PAN features \hfill (Eq. \ref{eq:mard})
    \STATE Update $\theta_S$ using $\mathcal{L}_{\text{det}} + \lambda\mathcal{L}_{\text{mard}}$
  \ENDFOR
  \STATE Update teacher: $\theta_T \leftarrow \mu \theta_T + (1-\mu)\theta_S$ \hfill (epoch-level EMA)
\ENDFOR
\end{algorithmic}
\end{algorithm}

\vspace{4pt}
\noindent \textbf{DHF: Dual-Head Pseudo-Label Fusion}
\noindent Let $\mathcal{P}^{o}$ and $\mathcal{P}^{m}$ denote the teacher predictions from the O2O and O2M heads on the weak view $x^w$, respectively.
Each prediction is a tuple $p=(b,s,c)$ where $b\in\mathbb{R}^{4}$ denotes box coordinates $(x_1,y_1,x_2,y_2)$, $s$ is the confidence, and $c$ is the class.

\noindent We first keep high-confidence O2O predictions as anchors:
\begin{equation}
\hat{Y}^{o} = \{p\in\mathcal{P}^{o} \mid s(p)\ge \tau_o\}.
\end{equation}

\noindent We then consider O2M candidates
$\mathcal{C}^{m}=\{p\in\mathcal{P}^{m}\mid s(p)\ge \tau_m\}$,
and add only those that have low overlap w.r.t. the O2O anchors:
\begin{equation}
\mathcal{E} = \left\{p\in \mathcal{C}^{m} \;\middle|\; \max_{q\in \hat{Y}^{o}} \mathrm{IoU}(b_p,b_q) \le \tau_{\text{no}} \right\}.
\label{eq:extras}
\end{equation}
This rule uses the O2O head as a high-precision prior and lets O2M contribute only non-redundant boxes, improving pseudo-label coverage while mitigating noisy supervision during self-training.
Because O2M predictions can contain duplicates among themselves, we apply class-wise NMS \emph{only} on the O2M extras:
\begin{equation}
\mathcal{E} \leftarrow \mathrm{NMS}(\mathcal{E},\tau_{\text{dup}}).
\end{equation}
\vspace{-2pt}
Finally, the fused pseudo-label set is
\begin{equation}
\label{eq:DHF}
\hat{Y}^w = \hat{Y}^{o} \cup \mathcal{E}.
\end{equation}


\vspace{-2pt}
\noindent \textbf{MARD: Multi-scale Adaptive Representation Diversification}
\label{sec:mard}

\noindent Let $\{F_\ell^S\}_{\ell\in\{3,4,5\}}$ be the three feature maps input to the YOLOv10 detection head (PAN outputs),
where $F_\ell \in \mathbb{R}^{B\times C_\ell \times H_\ell \times W_\ell}$.
We compute MARD on these features during the student forward pass on $x^s$.
Crucially, MARD gradients propagate through the PAN into the backbone, actively shaping feature learning during adaptation.  Unlike representation-learning methods that regularize \emph{global} embeddings\cite{bardes2021vicreg}, MARD constructs its feature vector set using \emph{detection structure}:

\begin{itemize}[topsep=-3pt, noitemsep]
    \item \textbf{Pseudo-box guided foreground feature vectors.}
    From the pseudo-labels $\hat{Y}^w$, we sample $K$ spatial locations \emph{inside each pseudo box} and extract the corresponding vectors from the detection feature map $F_\ell$.
    This ties MARD to \emph{object evidence} used by the detector.
    \item \textbf{Complementary background feature vectors.}
    We also sample $M$ locations from \emph{background} regions (valid, non-padded pixels outside pseudo boxes), encouraging object-background separability in the same feature space.
    \item \textbf{Noise control under SFOD.}
    We restrict sampling to the top-$K_b$ pseudo boxes with confidence above $\tau_b$.
    This improves robustness to pseudo-label noise, and the adaptive weighting in Eq.~\eqref{eq:lambda} further downweights MARD when pseudo-label quality is low.
\end{itemize}

\noindent YOLO-style detectors distribute predictions across feature pyramid levels (PAN) according to object size.
To keep MARD detection consistent, we assign each pseudo box $b$ to a feature level based on its size
$s(b)=\sqrt{w(b)h(b)}$ in pixels and stride-derived thresholds:
\begin{equation}
g(b)=
\begin{cases}
3 & s(b)\le \eta\cdot \mathrm{stride}_3,\\
4 & \eta\cdot \mathrm{stride}_3 < s(b)\le \eta\cdot \mathrm{stride}_4,\\
5 & s(b) > \eta\cdot \mathrm{stride}_4,
\end{cases}
\label{eq:levelassign}
\end{equation}
where $\eta$ is a constant and $\mathrm{stride}_\ell$ is the effective stride of level $\ell$.
We then sample foreground/background feature vectors \emph{per level} using only boxes assigned to that level.
This prevents MARD from mixing incompatible scales and directly targets the scale-specific features used for small/medium/large objects.

For each level $\ell$, we collect a feature vector matrix $Z_\ell \in \mathbb{R}^{N_\ell \times C_\ell}$, containing foreground and background feature vectors.
We then apply two complementary constraints:
(i) a \emph{variance} term to prevent feature collapse (ensuring each channel maintains a non-trivial spread), and
(ii) a \emph{covariance} term to reduce redundancy across channels. 
\begin{align}
\mathcal{L}_{\text{var}}(Z_\ell)
&=
\frac{1}{C}\sum_{c=1}^{C} \max\left(0,\gamma-\sqrt{\mathrm{Var}(Z_{\ell,:,c})+\epsilon}\right),
\label{eq:var}
\\
\mathcal{L}_{\text{cov}}(Z_\ell)
&=
\frac{1}{C(C-1)}\sum_{i\neq j}
\left(\mathrm{Cov}(\tilde{Z}_\ell)_{ij}\right)^2,
\label{eq:cov}
\end{align}
where $\tilde{Z}_\ell$ denotes channel-wise normalized tokens, $\gamma$ is a hyperparameter and $\epsilon$ is a small constant for stability.
While the variance and covariance terms share a functional form with collapse-prevention objectives used in self-supervised learning~\cite{bardes2021vicreg}, MARD differs in that it operates on \emph{multi-scale PAN features} rather than global embeddings, constructs feature vector sets via \emph{pseudo-box guided foreground and background sampling} tied to detection structure, and enforces \emph{scale-consistent} regularization via the level assignment in Eq.~\eqref{eq:levelassign}.

\noindent The MARD loss is defined as:
\begin{equation}
\mathcal{L}_{\text{mard}} = \sum_{\ell\in\{3,4,5\}}
\left(
\alpha\,\mathcal{L}_{\text{var}}(Z_\ell) + \beta\,\mathcal{L}_{\text{cov}}(Z_\ell)
\right).
\label{eq:mard}
\end{equation}

\noindent To avoid over-regularizing when pseudo-labels are unreliable, we use an adaptive weight
\begin{equation}
\lambda(t) = \lambda_0 \cdot \mathrm{ramp}(t) \cdot \mathrm{gate}(\bar{s}),
\label{eq:lambda}
\end{equation}

where $\mathrm{ramp}(t)$ linearly increases from $0$ to $1$ over a warmup period, and
$\mathrm{gate}(\bar{s})$ increases with the batch-average pseudo-label confidence $\bar{s}$ (clipped to $[0,1]$).
We also compute MARD periodically (every $I$ steps) to limit overhead.

\subsection{Training Objective and Inference}
\label{sec:objective}
Given the student outputs on $x^s$ and pseudo-labels $\hat{Y}^w$ mapped to strong-view $\hat{Y}^s$, we first calculate the standard YOLOv10 detection loss (box regression, classification, and distribution focal loss), computed over \emph{both} the O2O and O2M student heads to retain the dual-head training dynamics of YOLOv10:
\begin{equation}
\mathcal{L}_{\text{det}} = \mathcal{L}_{\text{box}} + \mathcal{L}_{\text{cls}} + \mathcal{L}_{\text{dfl}}.
\end{equation}

\noindent \textbf{Total loss.}
Our final optimization objective is:
\begin{equation}
\mathcal{L} = \mathcal{L}_{\text{det}} + \lambda(t)\,\mathcal{L}_{\text{mard}}.
\label{eq:total}
\end{equation}

\noindent \textbf{Inference-time efficiency.}
Dual-Head Fusion and MARD are used only during \emph{training-time} adaptation.
After adaptation, we deploy the student $f_{\theta_S}$ as a standard YOLOv10 detector, preserving \textbf{end-to-end NMS-free inference} and incurring \emph{no additional runtime cost}.

\begin{table*}[h]
\caption{%
  Comparison with state-of-the-art on \textbf{C2F} (Cityscapes $\to$ Foggy Cityscapes).
  \textbf{Bold} denotes the best SFOD result; \underline{underline} the second-best.
  Efficiency metrics are measured using each method's official public codebase.
  $^\dagger$FRANCK has no public implementation; its efficiency figures are adopted from DRU, which shares the same base detector. S\,=\,Source-only; UDAOD\,=\,Unsupervised DA OD; SFOD\,=\,Source-Free OD. All our results are averaged over three random seeds.
}
\centering
\vspace{4pt}
\resizebox{\textwidth}{!}{%
\renewcommand{\arraystretch}{1.2}
\begin{tabular}{%
  l                  
  l                  
  l                  
  c c c              
  cccccccc           
  c                  
}
\toprule
\textbf{Category} &
\textbf{Method} &
\textbf{Base} &
\textbf{Params (M)} &
\textbf{FPS} &
\textbf{Lat.\ (ms)} &
\textbf{prsn} &
\textbf{rider} &
\textbf{car} &
\textbf{truck} &
\textbf{bus} &
\textbf{train} &
\textbf{mcycle} &
\textbf{bicycle} &
\textbf{mAP} \\
\midrule

\multirow{3}{*}{\textbf{S}}
  & \multirow{3}{*}{Source only} & YOLOv10S & 7.2  & 233 & 4.3  & 41.0 & 46.1 & 49.6 & 12.0 & 27.0 &  4.9 & 17.1 & 39.0 & 29.6 \\
& & YOLOv10M & 15.4 & 105 & 9.5  & 43.6 & 47.8 & 50.0 & 12.8 & 29.8 &  5.0 & 20.1 & 40.2 & 31.2 \\
& & YOLOv10L & 24.4 &  67 & 15.0 & 47.1 & 52.4 & 58.5 & 18.6 & 35.1 & 10.8 & 24.5 & 44.8 & 36.5 \\
\midrule

\multirow{5}{*}{\textbf{UDAOD}}
  & AT \scriptsize{(CVPR'22)}       & FRCNN & NA & NA & NA & 43.7 & 54.1 & 62.3 & 31.9 & 54.4 & 49.3 & 35.2 & 47.9 & 47.4 \\
& MRT \scriptsize{(ICCV'23)}        & DETR  & NA & NA & NA & 52.8 & 51.7 & 68.7 & 35.9 & 58.1 & 54.5 & 41.0 & 47.1 & 51.2 \\
& CAT \scriptsize{(CVPR'24)}        & FRCNN & NA & NA & NA & 44.6 & 57.1 & 63.7 & 40.8 & 66.0 & 49.7 & 44.9 & 53.0 & 52.5 \\
& DATR \scriptsize{(TIP'25)}        & DETR  & NA & NA & NA & 61.6 & 60.4 & 74.3 & 35.7 & 60.3 & 35.4 & 43.6 & 55.9 & 53.4 \\
& SEEN-DA \scriptsize{(CVPR'25)}    & FRCNN & NA & NA & NA & 58.5 & 64.5 & 71.7 & 42.0 & 61.2 & 54.8 & 47.1 & 59.9 & 57.5 \\
\midrule

\multirow{12}{*}{\textbf{SFOD}}
  & IRG \scriptsize{(CVPR'23)}              & \multirow{4}{*}{FRCNN}    & 34.0  & 51 & 19.5 & 37.4 & 45.2 & 51.9 & 24.4 & 39.6 & 25.2 & 31.5 & 41.6 & 37.1 \\
& PETS \scriptsize{(ICCV'23)}               &                           & 44.2  & 48 & 20.6 & 42.0 & 48.7 & 56.3 & 19.3 & 39.3 &  5.5 & 34.2 & 41.6 & 35.9 \\
& Simple-SFOD \scriptsize{(ECCV'24)}        &                           & 43.8  & 42 & 23.8 & 40.9 & 48.0 & 58.9 & 29.6 & 51.9 & \underline{50.2} & 36.2 & 44.1 & 45.0 \\
& FALCON-SFOD \scriptsize{(CVPR'26)}        &                           & 43.8  & 42 & 23.8 & 41.0 & 48.3 & 58.7 & 33.6 & \underline{54.8} & \textbf{54.3} & 38.6 & 46.2 & 46.9 \\
\cline{2-15}
  & SF-YOLO \scriptsize{(ECCV'24 ws)}       & YOLOv5S                   &  7.2  & 172 &  5.8 & 48.6 & 51.8 & 64.5 & 24.7 & 48.1 & 37.6 & 23.1 & 44.8 & 42.9 \\
& SF-YOLO \scriptsize{(ECCV'24 ws)}         & YOLOv5L                   & 46.5  &  52 & 19.1 & \underline{55.5} & \underline{58.0} & \textbf{71.5} & \underline{36.6} & 53.7 & 46.1 & \textbf{40.6} & \underline{50.5} & \underline{51.6} \\
\cline{2-15}
  & DRU \scriptsize{(ECCV'24)}              & \multirow{3}{*}{Def-DETR} & 40.0  & 29 & 35.0 & 48.3 & 51.5 & 62.5 & 26.2 & 43.2 & 34.1 & 34.2 & 48.6 & 43.6 \\
& FRANCK \scriptsize{(TIP'25)}              &                           & 40.0$^\dagger$ & 29$^\dagger$ & 35$^\dagger$ & 48.1 & 49.3 & 60.6 & 33.9 & 48.2 & 36.9 & 34.0 & 47.9 & 44.9 \\
& VFM-SFOD \scriptsize{(AAAI'26)}          &                           & 41.0  & 27 & 37.0 & 48.8 & 51.0 & 62.8 & \underline{36.6} & 52.7 & 39.9 & 36.4 & 48.2 & 47.1 \\
\cline{2-15}
  \rowcolor{ourRow}
  &                                         & YOLOv10S & \textbf{7.2} & \textbf{233} & \textbf{4.3} & 47.6 & 52.6 & 64.8 & 25.9 & 46.0 & 41.5 & 30.4 & 45.2 & 44.3 \\
  \rowcolor{ourRow}
  & \textbf{RT-SFOD (Ours)}                 & YOLOv10M & \textbf{15.4} & \textbf{105} & \textbf{9.5} & 53.7 & 57.4 & 67.3 & 26.1 & 46.3 & 44.9 & 35.8 & 47.3 & 47.4 \\
  \rowcolor{ourRow}
  &                                         & YOLOv10L & \textbf{24.4} & \textbf{67} & \textbf{15} & \textbf{58.4} & \textbf{64.7} & \underline{71.0} & \textbf{38.5} & \textbf{57.0} & 49.5 & \underline{39.0} & \textbf{52.1} & \textbf{53.8} \\
\bottomrule
\end{tabular}%
}
\label{tab:c2f}
\vspace{-14pt}
\end{table*}
\section{Experiments}
\label{sec:experiments}
\vspace{-16pt}
\textbf{Datasets and Metrics.} Following existing works, we use five publicly available datasets covering four domain-shift scenarios: Cityscapes\cite{cityscapes}, Foggy Cityscapes\cite{foggy_cityscapes}, KITTI\cite{kitti}, Sim10k\cite{sim10k}, and BDD100k\cite{bdd100k}. Detailed description of datasets is provided in supplementary (Sec. S.4.2). Unlike existing SFOD methods, which only report target domain accuracy (mean average precision mAP at 0.5 IoU threshold), we report mAP along with \emph{efficiency metics} like throughout (FPS), number of parameters, and latency (ms/img). 
As noted in \cite{rf-detr}, many efficiency-focused methods \cite{rtdetr, sf-yolo, yolov10} report FP32 PyTorch accuracy but FP16 TensorRT efficiency, leading to an unfair comparison. Moreover, \cite{rf-detr} shows that naive FP16 quantization can significantly degrade performance. Therefore, we report all primary results in FP32 using PyTorch, and provide FP16 TensorRT results in the supplementary (Sec. S.8) for comparability.

\noindent \textbf{Implementation details.} The student is trained using SGD with an initial learning rate of 1e-4 and cosine annealing, gradient clipping at norm 10, and batch size 16 for 60 epochs. The teacher is updated via Exponential Moving Average (EMA) with momentum 0.999 at the end of each epoch. All experiments are conducted on a single NVIDIA RTX A6000 GPU. We use the same hyperparameters across all domain-shifts. Details about hyperparameters are included in the supplementary (Sec. S.4.1).

\subsection{Comparison with State-of-the-Art Methods}
\label{sec:sota}
Following existing works, we compare our method with (i)UDA-based object detection (UDAOD), which retains source data during adaptation \cite{at, mrt, cat, datr, seen-da}, and (ii) SFOD methods \cite{irg, pets, simple-sfod, dru, franck, falcon-sfod, beyond_boundaries, sf-yolo}.

\begin{wraptable}[16]{r}{0.42\textwidth}
\vspace{-20pt}
\caption{S2C and K2C comparison. Efficiency metrics are identical to those in Table \ref{tab:c2f}}
\vspace{-4pt}
\centering
\scriptsize
\resizebox{0.4\textwidth}{!}{%
\renewcommand{\arraystretch}{1.1}
\begin{tabular}{llccc}
\toprule
\textbf{Cat.} & \textbf{Method} & \textbf{Base} & \textbf{S2C} & \textbf{K2C} \\
\midrule
\multirow{3}{*}{\textbf{S}}
  & \multirow{3}{*}{Source only} & YOLOv10S & 51.2 & 37.4 \\
& & YOLOv10M & 53.5 & 40.3 \\
& & YOLOv10L & 56.5 & 45.0 \\
\midrule
\multirow{2}{*}{\textbf{UDA}}
  & DATR\cite{datr}      & DETR  & 64.3 & 53.3 \\
& SEEN-DA\cite{seen-da} & FRCNN & 66.8 & 67.1 \\
\midrule
\multirow{10}{*}{\textbf{SFOD}}
  & IRG\cite{irg}              & \multirow{5}{*}{FRCNN} & 45.2 & 46.9 \\
& PETS\cite{pets}              &                     & 57.8 & 47.0 \\
& LPLD\cite{lpld}              &                     & 49.4 & 51.3 \\
& Simple\cite{simple-sfod}     &                     & 55.4 & 46.2 \\
& FALCON\cite{falcon-sfod}     &                     & 58.8 & 50.1 \\
\cline{2-5}
& SF-YOLO\cite{sf-yolo} & YOLOv5S & 57.7 & 50.6 \\
& SF-YOLO\cite{sf-yolo} & YOLOv5L & \underline{69.8} & \textbf{63.7} \\
\cline{2-5}
& DRU\cite{dru}               & \multirow{3}{*}{DefDETR} & 58.7 & 45.1 \\
& FRANCK\cite{franck}          &                     & 63.1 & 48.5 \\
& VFM\cite{beyond_boundaries} &                     & 67.4 & 54.7 \\
\cline{2-5}
\rowcolor{ourRow}
&  & YOLOv10S & 59.8 & 52.3 \\
\rowcolor{ourRow}
& \textbf{RT-SFOD (Ours)} & YOLOv10M & 66.4 & 54.2 \\
\rowcolor{ourRow}
& & YOLOv10L & \textbf{71.2} & \underline{60.9} \\
\bottomrule
\end{tabular}%
}
\label{tab:s2c_k2c}
\vspace{-2pt}
\end{wraptable}

\noindent\textbf{Cityscapes $\to$ Foggy Cityscapes (C2F).}
Table~\ref{tab:c2f} presents results on the most widely used SFOD benchmark. Our RT-SFOD-L achieves \textbf{53.8 mAP}, surpassing the previous best SF-YOLO-L (51.6) by {+2.2} mAP and remaining competitive with UDAOD methods (MRT: 51.2, DATR: 53.4) under the stricter source-free constraint. Our RT-SFOD-M (47.4 mAP) outperforms most Faster R-CNN and Deformable DETR based SFOD approaches. Compared to SF-YOLO-L, RT-SFOD-L is {1.3$\times$} faster (67 vs.\ 52 FPS) and {1.9$\times$} smaller (24.4M vs.\ 46.5M parameters). Against Deformable DETR-based methods (DRU, FRANCK, VFM-SFOD), which cluster around 27--29~FPS and 40--41M parameters, RT-SFOD-M matches their best accuracy at \textbf{3.6$\times$ higher throughput} and \textbf{2.6$\times$ fewer parameters}. Our smallest variant, RT-SFOD-S (44.3 mAP), is on par with DRU (43.6) and FRANCK (44.9) while running at \textbf{233 FPS} with only \textbf{7.2M} parameters. These trade-offs are visualized in Fig.~\ref{fig:teaser}.

\vspace{2pt}
\noindent\textbf{Sim10k$\to$Cityscapes (S2C) and KITTI $\to$Cityscapes (K2C).}
Following prior works, we report AP(on Car) for S2C and K2C. On S2C, our RT-SFOD-L achieves \textbf{71.2 mAP}, outperforming SF-YOLO-L (69.8) by {+1.4} and VFM-SFOD (67.4) by {+3.8}. On K2C, it reaches \underline{60.9 mAP}, second only to SF-YOLO-L (63.7), while using \textbf{1.9$\times$ fewer parameters} and \textbf{1.3$\times$ higher throughput}. We attribute the gap to SF-YOLO-L to model capacity: SF-YOLO-L uses 1.9$\times$ more parameters (46.5M vs.\ 24.4M), and at matched scale (both 7.2M), RT-SFOD-S outperforms SF-YOLO-S (52.3 vs.\ 50.6). Moreover, RT-SFOD-S/M outperform all five Faster R-CNN methods and two of three DETR-based methods on K2C while being faster and smaller, indicating that this camera-shift scenario particularly benefits from additional model capacity. The S$\to$M$\to$L scaling yields 59.8$\to$66.4$\to$71.2 on S2C and 52.3$\to$54.2$\to$60.9 on K2C, demonstrating consistent gains as model size increases.

\vspace{2pt}
\noindent\textbf{Cityscapes $\to$ BDD100k (C2B).}
Table~\ref{tab:c2b} presents a challenging large-scale benchmark spanning diverse cities, weather conditions, and times of day. Our RT-SFOD-L achieves \textbf{46.5 mAP}, exceeding VFM-SFOD (43.0) by {+3.5} and FALCON-SFOD (36.9) by {+9.6}, and also beating both UDAOD baselines. Even RT-SFOD-S (41.1~mAP) is competitive with heavyweight Def-DETR methods such as FRANCK (40.7). The consistent gains across C2F (weather), C2B (large-scale diversity), S2C (rendering gap), and K2C (camera shift) indicate that our modules generalize across domain gap characteristics.

\begin{table}[t]
\caption{Comparison with state-of-the-art on C2B (Cityscapes $\to$ BDD100k). Efficiency metrics are identical to those in Table \ref{tab:c2f}.}
\centering
\resizebox{\columnwidth}{!}{%
\renewcommand{\arraystretch}{1.2}
\begin{tabular}{llccccccccc}
\toprule
\textbf{Category} & \textbf{Method} & \textbf{Base} & \textbf{truck} & \textbf{car} & \textbf{rider} & \textbf{person} & \textbf{motor} & \textbf{bicycle} & \textbf{bus} & \textbf{mAP} \\
\midrule
\multirow{3}{*}{\textbf{S}}
  & \multirow{3}{*}{Source only} & YOLOv10S & 17.5 & 59.6 & 29.1 & 42.4 & 14.6 & 19.3 & 20 & 28.9 \\
& & YOLOv10M & 18.8 & 61.8 & 32 & 44.5 & 15.8 & 22.6 & 21.5 & 31 \\
& & YOLOv10L & 23.1 & 68.3 & 35.4 & 49.2 & 17.8 & 24 & 25.3 & 34.7 \\
\midrule
\multirow{2}{*}{\textbf{UDA}}
  & MRT\cite{mrt} \venue{ICCV'23}  & Def-DETR & 24.7 & 63.7 & 30.9 & 48.4 & 20.2 & 22.6 & 25.5 & 33.7 \\
& DATR\cite{datr} \venue{TIP'25}   & Def-DETR & 26.9 & 73.4 & 42.8 & 58.5 & 24.2 & 37.3 & 39.9 & 43.3 \\
\midrule
\multirow{10}{*}{\textbf{SFOD}}
  & IRG\cite{irg} \venue{CVPR'23}              & \multirow{4}{*}{FRCNN} & 31.4 & 59.7 & 32.8 & 39.9 & 16.7 & 26.9 & 21.5 & 32.7 \\
& PETS\cite{pets} \venue{ICCV'23}              & & 19.3 & 62.4 & 34.5 & 42.6 & 17.0 & 26.3 & 16.9 & 31.3 \\
& Simple-SFOD\cite{simple-sfod} \venue{ECCV'24} & & 32.0 & 60.0 & 33.4 & 40.2 & 19.7 & 29.9 & 24.9 & 34.3 \\
& FALCON-SFOD\cite{falcon-sfod}                & & 32.6 & 59.8 & 34.0 & 40.0 & 25.7 & \textbf{35.7} & 30.5 & 36.9 \\
\cline{2-11}
& DRU\cite{dru} \venue{ECCV'24}              & \multirow{3}{*}{Def-DETR} & 27.1 & 62.7 & 36.9 & 45.8 & 22.7 & \underline{32.5} & 28.1 & 36.6 \\
& FRANCK\cite{franck}                        & & 29.5 & 65.6 & 43.3 & 55.0 & 30.0 & 28.0 & 33.6 & 40.7 \\
& VFM-SFOD\cite{beyond_boundaries}           & & 33.2 & 72.3 & 44.2 & 54.9 & 32.9 & 29.0 & 34.8 & 43.0 \\
\cline{2-11}
\rowcolor{ourRow}
& & YOLOv10S & 28.8 & 71.6 & 44.8 & 53.8 & 28.7 & 26.5 & 33.6 & 41.1 \\
\rowcolor{ourRow}
& \textbf{RT-SFOD (Ours)} & YOLOv10M & \underline{33.7} & \underline{75.9} & \underline{47.3} & \underline{57.5} & \underline{33} & 28.6 & \underline{35.8} & \underline{44.5} \\
\rowcolor{ourRow}
& & YOLOv10L & \textbf{36.4} & \textbf{77.0} & \textbf{49.3} & \textbf{58} & \textbf{34.1} & 31.4 & \textbf{39.6} & \textbf{46.5} \\
\bottomrule
\end{tabular}%
}
\label{tab:c2b}
\vspace{-20pt}
\end{table}

\subsection{Ablation Study}
\label{sec:ablation}
\vspace{-25pt}
\begin{table}[h]
\caption{Ablation study of the proposed components with RT-SFOD-M across domain-shifts C2F, S2C, K2C.}
\centering
\resizebox{\columnwidth}{!}{%
\begin{tabular}{l cc ccccccccc c c}
\toprule
\multirow{2}{*}{\textbf{Module}} & \multicolumn{2}{c}{\textbf{Components}} & \multicolumn{9}{c}{\textbf{C2F}} & \textbf{S2C} & \textbf{K2C} \\
\cmidrule(lr){2-3} \cmidrule(lr){4-12} \cmidrule(lr){13-13} \cmidrule(lr){14-14}
& \textbf{DHF} & \textbf{MARD} & \textbf{prsn} & \textbf{rider} & \textbf{car} & \textbf{truck} & \textbf{bus} & \textbf{train} & \textbf{mcycle} & \textbf{bicycle} & \textbf{mAP} & \textbf{AP} & \textbf{AP} \\
\midrule
Source-trained        & $\times$ & $\times$ & 45.6 & 48.8 & 50   & 7.48 & 27   & 3    & 20.1 & 42.2 & 30.5 & 52.5 & 40.3 \\
AdaBN                 & $\times$ & $\times$ & 50.9 & 51.4 & 57.9 & 15.7 & 35.8 & 15.5 & 26.6 & 45.1 & 37.4 & 54.2 & 45.2 \\
MT (O2M)              & $\times$ & $\times$ & 50.3 & 54.5 & 63.7 & 24.3 & 45.0 & 41.1 & 33.5 & 45.3 & 44.7 & 61.4 & 49.9 \\
MT (O2O)              & $\times$ & $\times$ & 51.5 & 55.0 & 63.8 & 24.9 & 45.2 & 41.7 & 33.8 & 45.9 & 45.2 & 62.1 & 50.6 \\
MT (O2O) + DHF        & \checkmark & $\times$ & 52.9 & 56.4 & 67.0 & 25.5 & 45.6 & 44.3 & 35.2 & 46.6 & 46.7 & 65.1 & 53.0 \\
MT (O2O) + MARD       & $\times$ & \checkmark & 51.8 & 55.7 & 66.2 & 25.2 & 45.9 & 44.7 & 35.0 & 46.2 & 46.3 & 65.3 & 52.5 \\
MT (O2O) + DHF + MARD & \checkmark & \checkmark & 53.7 & 57.4 & 67.3 & 26.1 & 46.3 & 44.9 & 35.8 & 47.3 & \textbf{47.4} & \textbf{66.4} & \textbf{54.2} \\
\bottomrule
\end{tabular}%
}
\label{tab:ablation}
\vspace{-16pt}
\end{table}

Table \ref{tab:ablation} presents a component-wise ablation with RT-SFOD-M across three benchmarks. Starting from the source-trained model (30.5/52.5/40.3 mAP on C2F/ S2C/K2C), AdaBN provides a better starting point for the teacher by aligning batch normalization statistics before training. Adding mean-teacher (MT) self-training yields further gains: MT with one-to-many (O2M) pseudo-labels achieves 44.7 mAP on C2F, while MT with one-to-one (O2O) obtains 45.2. The O2O head's slight advantage is consistent with our precision-recall analysis in Fig. \ref{fig:pl_analysis}: while O2M with standard post-processing provides broader coverage (recall 0.469 vs.\ 0.446), it introduces additional noise (precision 0.847 vs.\ 0.923) that can compound during iterative self-training. However, despite these gains, the vanilla MT framework remains sub-optimal.

\noindent\textbf{Effect of DHF.}
Adding Dual-Head Fusion to MT~(O2O) yields consistent gains: {+1.5} on C2F (45.2 $\to$ 46.7), {+3.0} on S2C (62.1 $\to$ 65.1), and {+2.4} on K2C (50.6 $\to$ 53.0), with the largest per-class gains on car (+3.2) and train (+2.6). 

\noindent\textbf{Effect of MARD.}
Adding MARD loss (w/o DHF) yields {+1.1} on C2F (45.2 $\to$ 46.3), {+3.2} on S2C (62.1 $\to$ 65.3), and {+1.9} on K2C (50.6 $\to$ 52.5). As shown in Fig.~\ref{fig:feat_analysis}, MARD recovers 76--84\% of the effective rank lost to domain-shift at each PAN-feature level (vs.\ only 26--48\% without it), restoring feature discriminability.

\noindent\textbf{Complementarity of DHF and MARD.}
Combining both modules achieves the best results: \textbf{47.4}/\textbf{66.4}/\textbf{54.2} on C2F/S2C/K2C, with total gains of {+2.2/ +4.3/+3.6} over MT~(O2O). The combined improvement exceeds either module in isolation, confirming they address \emph{complementary bottlenecks}: DHF improves the supervisory signal while MARD improves learned representations.

\vspace{-12pt}
\begin{table}[h]
\caption{Ablation studies on RT-SFOD-M. Left: MARD sub-objectives. Right: Pseudo-label strategy comparison.}
\centering
\begin{subtable}[t]{0.48\textwidth}
\vspace{-6pt}
\caption{Ablation of MARD components. Baseline: MT (O2O) + DHF.}
\centering
\resizebox{\textwidth}{0.06\textheight}{%
\begin{tabular}{ccccc}
\toprule
\textbf{Variance} & \textbf{Covariance} & \textbf{C2F} & \textbf{S2C} & \textbf{K2C} \\
\midrule
$\times$    & $\times$    & 46.7 & 65.1 & 53.0 \\
\checkmark  & $\times$    & 47.2 & 65.9 & 53.8 \\
$\times$    & \checkmark  & 47.0 & 65.7 & 53.5 \\
\checkmark  & \checkmark  & \textbf{47.4} & \textbf{66.4} & \textbf{54.2} \\
\bottomrule
\end{tabular}%
}
\label{tab:mard_subcomp}
\end{subtable}
\hfill
\begin{subtable}[t]{0.48\textwidth}
\vspace{-6pt}
\caption{Pseudo-label strategy comparison (RT-SFOD-M, C2F).}
\centering
\resizebox{\textwidth}{0.06\textheight}{%
\begin{tabular}{lcccc}
\toprule
\textbf{Strategy} & \textbf{Prec} & \textbf{Recall} & \textbf{F1} & \textbf{mAP} \\
\midrule
O2O only             & 0.923 & 0.446 & 0.602 & 45.2 \\
O2M+NMS              & 0.847 & 0.469 & 0.603 & 45.2 \\
O2O $\cup$ O2M+NMS   & 0.834 & 0.512 & 0.634 & 45.5 \\
DHF (Ours)           & 0.919 & 0.525 & \textbf{0.668} & \textbf{46.7} \\
\bottomrule
\end{tabular}%
}
\label{tab:dhf_ablation}
\end{subtable}
\label{tab:combined_ablation}
\vspace{-28pt}
\end{table}

\subsection{Analysis}
\label{sec:analysis}
\noindent\textbf{MARD sub-component analysis.}
Table~\ref{tab:mard_subcomp} ablates the two MARD components using MT~(O2O)~+~DHF as the fixed baseline. The variance term (preventing channel collapse) provides the larger individual contribution (+0.5/+0.8/+0.8 on C2F/S2C/K2C), consistent with channel collapse being the primary mode of rank degradation in Fig.~\ref{fig:feat_analysis}. The covariance term (reducing inter-channel redundancy) provides a complementary gain (+0.3/+0.6/+0.5). Combining both yields the full MARD improvement (+0.7/+1.3/+1.2), confirming they address distinct failure modes.

\noindent\textbf{Pseudo-label fusion strategy comparison.}
Table~\ref{tab:dhf_ablation} compares four pseudo-label strategies on label-quality metrics and downstream mAP. O2O yields high precision (0.923) but low recall (0.446), while O2M + NMS provides broader coverage (0.469) at lower precision (0.847), giving a comparable F1 (0.603). Directly merging O2O with O2M + NMS improves recall (0.512) but reduces precision (0.834), yielding a suboptimal F1 (0.634). Our DHF uses O2O boxes as high-confidence anchors and selectively adds non-redundant, high-scoring O2M boxes with low overlap w.r.t. O2O anchors, retaining near-O2O precision (0.919) while boosting recall to 0.525, achieving the best F1 (0.668) and mAP (46.7). This confirms that pseudo-label selection must balance precision and recall in self-training: the direct union improves recall but reduces mAP, as noisy pseudo-labels can compound across iterations.

\begin{wrapfigure}[18]{r}{0.54\textwidth}
    \centering
    \vspace{-20pt}
    \includegraphics[width=0.55\textwidth]{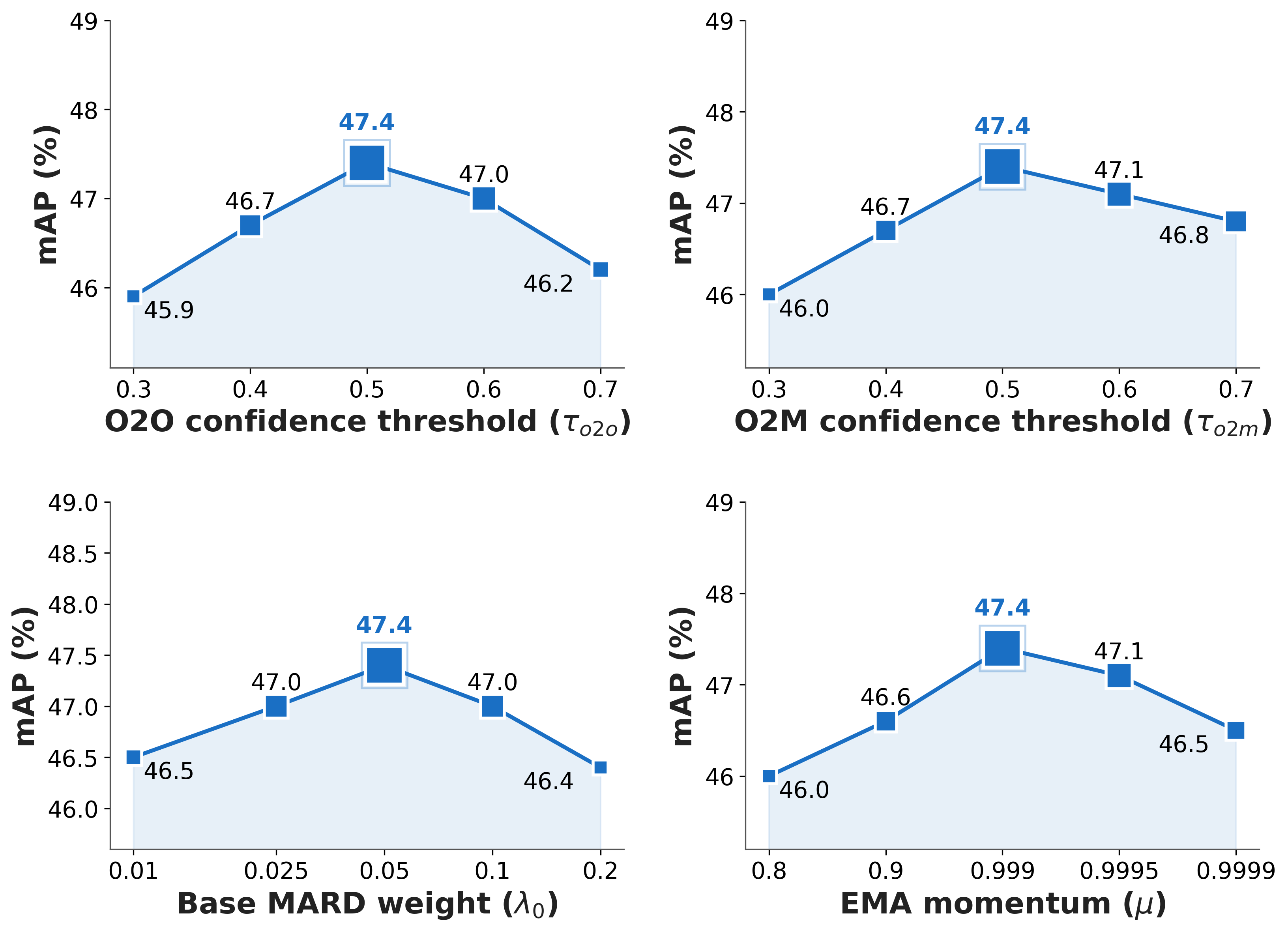}
    \vspace{-10pt}
    \caption{\textbf{Hyperparameter sensitivity} on C2F (RT-SFOD-M). mAP (\%) vs.\ O2O threshold $\tau_{o2o}$, O2M threshold $\tau_{o2m}$, MARD weight $\lambda_0$, and EMA momentum $\mu$. Variation is $\leq$1.5 mAP in all cases.}
    \label{fig:hyperparam}
\end{wrapfigure}

\noindent\textbf{Hyperparameter sensitivity.}
Fig.~\ref{fig:hyperparam} shows C2F mAP of RT-SFOD-M as a function of four key hyperparameters, with performance stable across all tested ranges ($\leq$1.5 mAP variation). Both confidence thresholds ($\tau_{o2o}$, $\tau_{o2m}$) peak at 0.5; the $\tau_{o2m}$ curve is notably flatter (46.0--47.4), indicating DHF's IoU-based deduplication provides robustness against the O2M threshold choice. The optimal MARD weight is $\lambda_0{=}0.05$, with performance degrading gracefully at both extremes; the effective weight is further modulated by a linear warmup and confidence gate (Sec.~\ref{sec:mard}). EMA momentum $\mu{=}0.999$ balances teacher stability and responsiveness. No hyperparameter requires per-benchmark tuning, and the same values transfer across detectors and scales, likely because they depend on relative statistics (calibrated confidence, geometric IoU, and feature distributions) rather than architecture-specific factors. The full set of values used is: $\tau_{o2o}{=}\tau_{o2m}{=}0.5$, $\tau_{no}{=}0.2$, $\tau_{dup}{=}0.7$, $\lambda_0{=}0.05$, $K{=}8$, $M{=}128$, $\mu{=}0.999$; full sensitivity analysis is in the supplementary (Sec. S.4.1).

\begin{wrapfigure}[18]{r}{0.5\linewidth}
    \centering
    \vspace{-20pt}    
    \includegraphics[width=\linewidth]{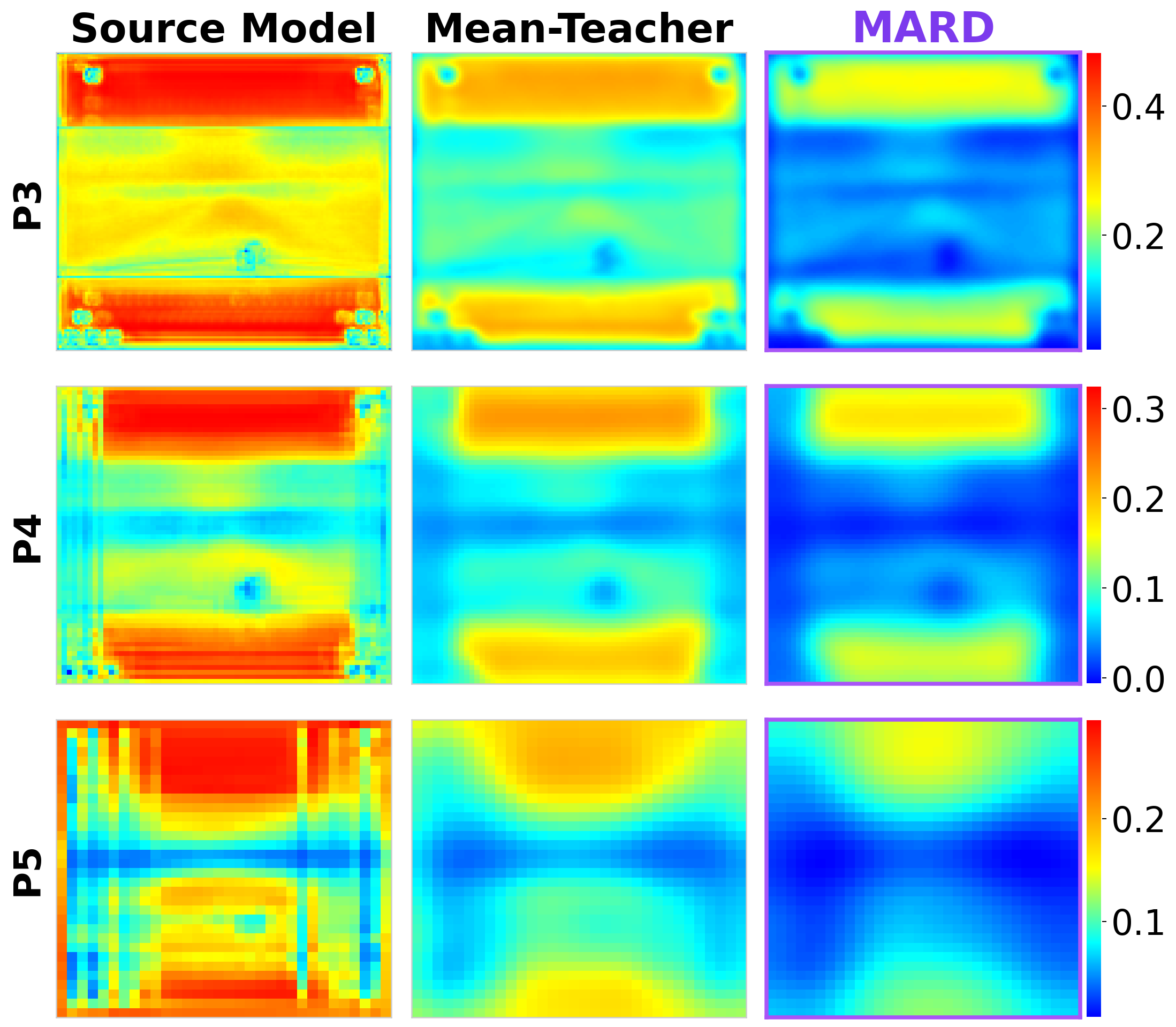}
    \caption{\footnotesize The average cosine similarity of each anchor point’s extracted features with all others across PAN scales(P3, P4, P5) on Foggy Cityscapes validation set.}
    \label{fig:cosine_similarity}
\end{wrapfigure}

\noindent\textbf{Feature Discriminability Analysis.}
Fig.~\ref{fig:cosine_similarity} illustrates the average cosine similarity between each anchor point’s extracted features and all others across PAN scales (P3, P4, and P5) on the Foggy Cityscapes validation set. We compare the Source model, Vanilla Mean-Teacher, and the proposed MARD approach. Notably, MARD consistently yields lower cosine similarity across all scales, indicating more structured feature representations and enhanced discriminability.

\noindent\textbf{Generality across Dual-Head Detectors.}
While our main experiments use YOLOv10, RT-SFOD can be applied to any NMS-free dual-head detector. DHF requires only a precise O2O head and a dense, high-recall O2M head, and MARD applies to any multi-scale feature pyramid. Table \ref{tab:generality} presents the analysis of RT-SFOD applied with the same hyperparameters to three additional dual-head detectors on C2F.

\begin{wraptable}[10]{r}{0.48\textwidth}
\vspace{-20pt}
\caption{Generality of RT-SFOD across dual-head detectors on C2F domain-shift.}
\vspace{-4pt}
\centering
\scriptsize
\setlength{\tabcolsep}{4pt}
\renewcommand{\arraystretch}{1.2}
\begin{tabular}{lcc}
\toprule
\textbf{Detector} & \textbf{MT (O2O)} & \textbf{RT-SFOD} \\
\midrule
YOLOv26S \cite{yolov26} & 42.1 & \textbf{44.6} \scriptsize{(+2.5)} \\
YOLOv26M \cite{yolov26} & 46.0 & \textbf{48.8} \scriptsize{(+2.8)} \\
YOLOv26L \cite{yolov26} & 50.6 & \textbf{52.7} \scriptsize{(+2.1)} \\
MS-DETR \cite{ms-detr}  & 42.3 & \textbf{44.3} \scriptsize{(+2.0)} \\
Mr.\,DETR \cite{mr-detr} & 43.8 & \textbf{46.1} \scriptsize{(+2.3)} \\
\bottomrule
\end{tabular}
\label{tab:generality}
\vspace{-8pt}
\end{wraptable}

As shown in Table~\ref{tab:generality}, RT-SFOD improves MT~(O2O) consistently across YOLOv26S/M/L \cite{yolov26} (+2.5/+2.8/+2.1), MS-DETR \cite{ms-detr} (+2.0), and Mr.\,DETR \cite{mr-detr} (+2.3), matching the gains observed on YOLOv10. This confirms that the benefits of DHF and MARD are properties of the dual-head paradigm rather than of a specific architecture. For single-head detectors, DHF has no second candidate set and reduces to confidence-threshold pseudo-labeling as in SF-YOLO \cite{sf-yolo}, while MARD remains applicable whenever pseudo-labels and multi-scale features are available.

\vspace{-8pt}
\section{Conclusion}
\vspace{-6pt}
\label{sec:conclusion}
We present \textbf{RT-SFOD}, a source-free object detection framework for NMS-free dual-head detectors, instantiated primarily on YOLOv10, that improves the accuracy-speed-model size trade-off for domain-adaptive detection. We identify two challenges in directly applying mean-teacher self-training to NMS-free dual-head detectors: (i) suboptimal pseudo-label supervision from unconditioned use of O2O and O2M predictions, and (ii) feature rank degradation under domain-shift. To address these, we propose \textbf{Dual-Head Fusion}, which uses high-precision O2O predictions as anchors and selectively incorporates non-redundant O2M candidates with low overlap relative to O2O anchors to increase recall while maintaining near-O2O precision. We also introduce \textbf{Multi-scale Adaptive Representation Diversification}, which enforces structured variance and covariance constraints on PAN features, recovering 76--84\% of rank lost under domain-shift compared to 26--48\% without it. Across domain-shift benchmarks, RT-SFOD improves mAP by 1.4--3.5\% over prior state-of-the-art methods, while the 1.3$\times$ higher throughput and smaller parameter count are inherited from the efficient YOLO backbone. We also demonstrate that our modules generalize across other dual-head detectors. Failure-case analysis and limitations are provided in the supplementary.

\section*{Acknowledgements} 
We gratefully acknowledge the support of the Ministry of Electronics and Information Technology and the Ministry of Education, Government of India, as well as IIT Hyderabad (India) and MBZUAI (Abu Dhabi) for their support of this project. We also sincerely thank the anonymous reviewers, area chairs, and program chairs for their valuable feedback, which helped improve the quality and presentation of the paper.

\bibliographystyle{splncs04}
\bibliography{main}

\appendix
\setcounter{page}{0}
\pagenumbering{arabic}

\setcounter{section}{0}
\setcounter{table}{0}
\setcounter{figure}{0}
\setcounter{equation}{0}

\renewcommand{\thesection}{S.\arabic{section}}
\renewcommand{\thetable}{S.\arabic{table}}
\renewcommand{\thefigure}{S.\arabic{figure}}
\renewcommand{\theequation}{S.\arabic{equation}}

\renewcommand{\theHsection}{S.\arabic{section}}
\renewcommand{\theHfigure}{S.\arabic{figure}}
\renewcommand{\theHtable}{S.\arabic{table}}
\renewcommand{\theHequation}{S.\arabic{equation}}

\newcommand{\ToCEntry}[3]{%
  \ifcase#1
    \noindent\ref{#3}. #2\hspace{1.5em}\dotfill\hspace{1.5em}\pageref{#3} \\ 
  \or
    \noindent\hspace*{2em}\ref{#3}. #2\hspace{1.5em}\dotfill\hspace{1.5em}\pageref{#3} \\ 
  \else
    \noindent\ref{#3}. #2\hspace{1.5em}\dotfill\hspace{1.5em}\pageref{#3} \\ 
  \fi
}

\title{Supplementary for \break Real-Time Source-Free Object Detection} 

\titlerunning{Real-Time Source-Free Object Detection}

\author{Sairam VCR\inst{1} \and
Varun Gopal\inst{1} \and
Poornima Jain\inst{1} \and \\
Vineeth N Balasubramanian\inst{1,2} \and
Muhammad Haris Khan\inst{3}}

\authorrunning{Sairam VCR et al.}

\institute{$^1$IIT Hyderabad, India \quad $^2$Microsoft Research \quad $^3$MBZUAI \\
\email{\{ai20resch13001@,co22btech11015@,ai24resch11002@,\\ vineethnb@cse.\}iith.ac.in}, \email{muhammad.haris@mbzuai.ac.ae}}

\maketitle
\setcounter{page}{1}

\section*{Contents}
\ToCEntry{0}{Feature Discriminability Analysis}{sec:heat_maps_supply}
\ToCEntry{0}{Qualitative Results}{sec:qual_res_supply}
\ToCEntry{1}{RT-SFOD is Robust to Partial Visibility}{sec:partial_visibility}
\ToCEntry{0}{Pseudo-label Quality Over Training Epochs}{sec:pl_quality}
\ToCEntry{0}{Reproducibility}{sec:reproducibility_supply}
\ToCEntry{1}{Hyperparameter Analysis}{subsec:hyperparams}
\ToCEntry{1}{Dataset Details}{subsec:dataset_details_supply}
\ToCEntry{1}{Details of Data Augmentation Strategies}{subsec:details_data_aug_supply}
\ToCEntry{0}{Failure-case Analysis}{sec:failure_case_supply}
\ToCEntry{0}{Teacher EMA Update Variations}{sec:ema_variation}
\ToCEntry{0}{Analysis of MARD Module}{sec:mard_analysis}
\ToCEntry{0}{TensorRT FP16 Efficiency Evaluation}{sec:trt_fp16}
\ToCEntry{0}{Training Overhead of Proposed Modules}{sec:training_overhead}
\ToCEntry{0}{Additional Analyses and Clarifications}{sec:additional}
\ToCEntry{1}{Generality Across Dual-Head Detectors}{sec:supp_generality}
\ToCEntry{1}{Robustness to Cold-Start and Noisy MARD Sampling}{sec:supp_robustness}
\ToCEntry{1}{Hyperparameter Selection and Transferability}{sec:supp_hyper}
\ToCEntry{1}{Limitations}{sec:supp_limitations}

\section{Feature Discriminability Analysis}
\label{sec:heat_maps_supply}
\begin{figure}[t]
  \centering
  \subfloat[Sim10k $\to$ Cityscapes]{\includegraphics[width=0.48\linewidth]{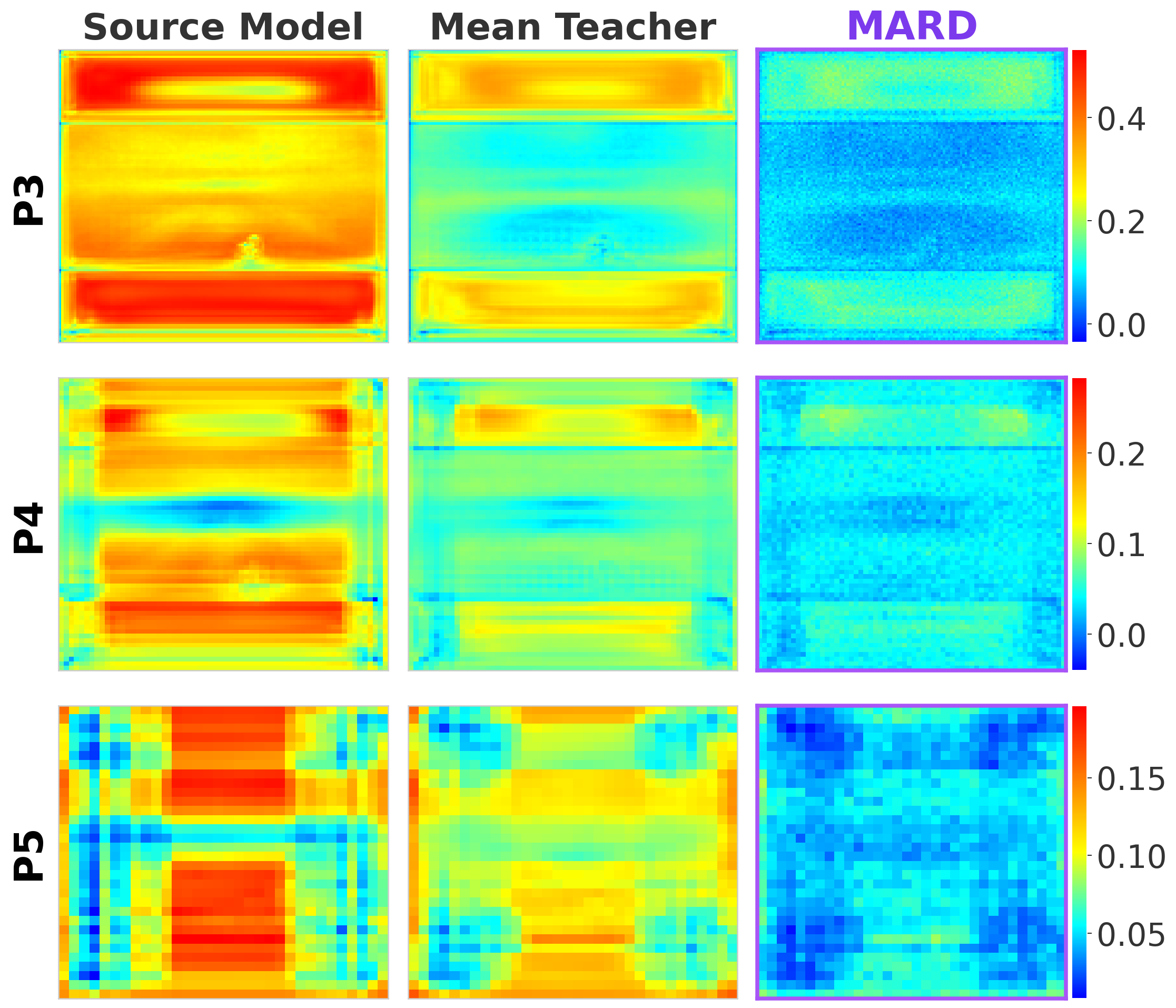} \label{fig:s2c_heatmaps}}
  \hfill
  \subfloat[KITTI $\to$ Cityscapes]{\includegraphics[width=0.48\linewidth]{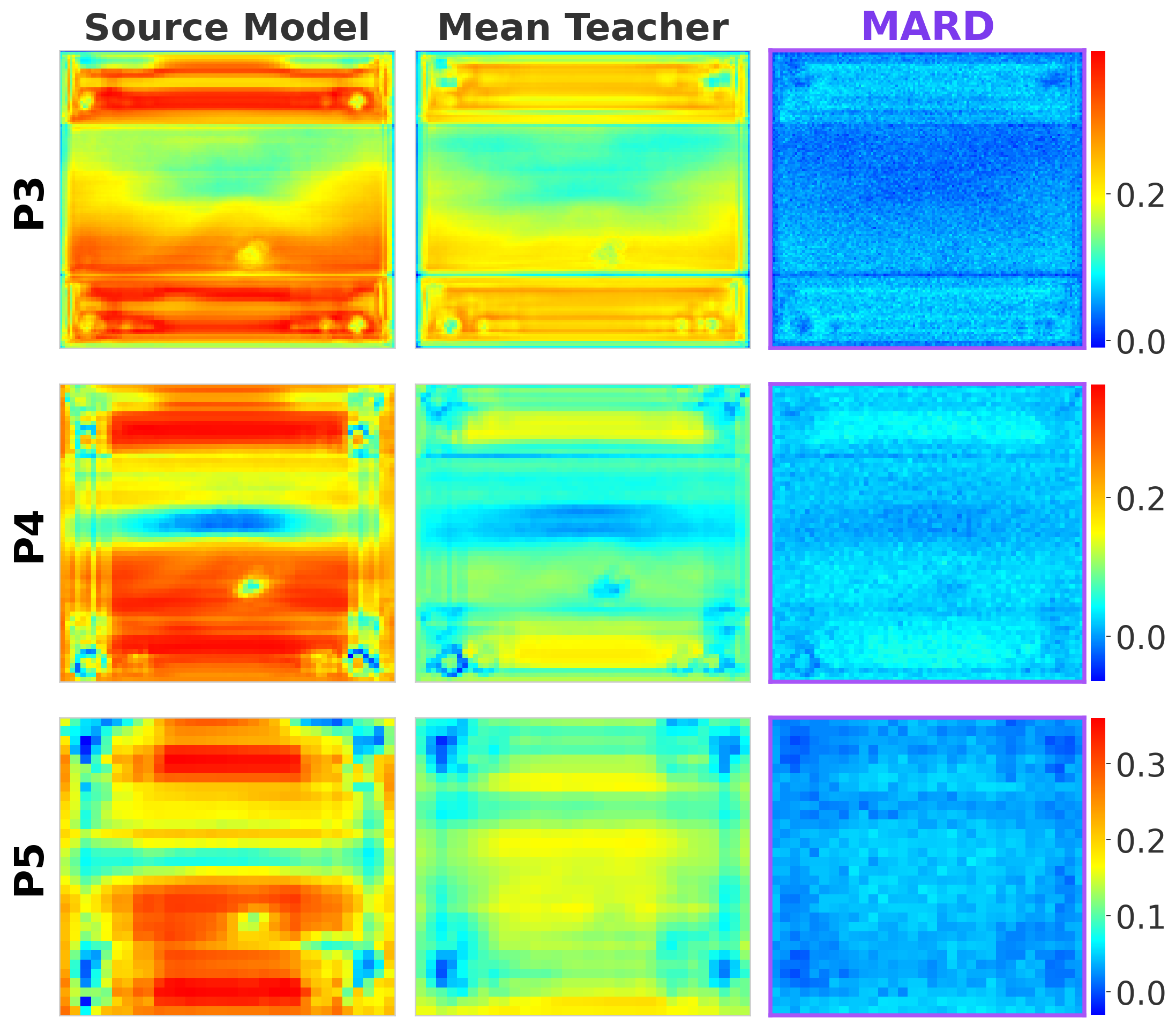} \label{fig:k2c_heatmaps}}
  \caption{The average cosine similarity (lower is better) of each anchor point’s extracted features with all others across feature-scales (P3, P4, P5). Across domain-shifts and feature-scales, MARD consistently achieves the \textit{lowest} cosine similarity.}
  \label{fig:cosine_similarity_supply}
  \vspace{-8pt}
\end{figure}

Similar to Fig. 5 in the main paper, we visualize the pairwise cosine similarity (lower is better) between each anchor point’s extracted features and all other anchors across feature-scales P3/P4/P5, for two additional domain shifts: Sim10k $\to$ Cityscapes and KITTI $\to$ Cityscapes (Fig. \ref{fig:cosine_similarity_supply}). Across domain-shifts and all feature levels, the source model exhibits highly correlated (less diverse) representations, and Mean Teacher provides only a partial reduction in similarity. In contrast, \textbf{MARD} consistently yields the \textbf{lowest cosine similarity}, indicating stronger representation discriminability and reduced feature collapse under domain-shift.

\section{Qualitative Results}
\label{sec:qual_res_supply}
\begin{figure}[t]
    \centering
    \includegraphics[width=\linewidth]{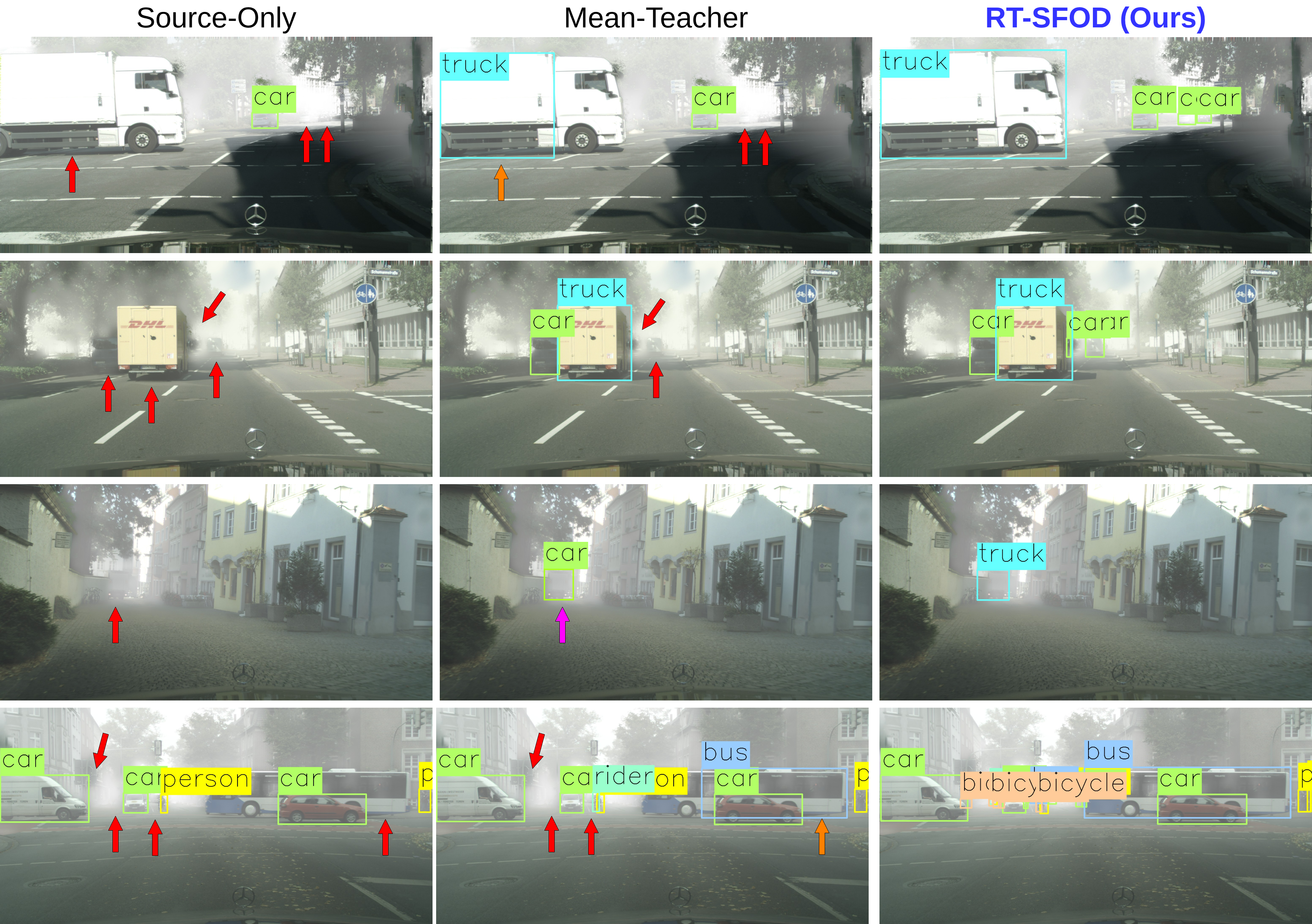}
    \caption{Qualitative comparison on the Foggy Cityscapes validation set between \textit{Source-Only}, \textit{Vanilla Mean-Teacher (o2o)}, and our proposed \textbf{RT-SFOD}. The \textcolor{red}{red}, \textcolor{orange}{orange}, and \textcolor{magenta}{pink} arrows indicate missed detections, inaccurate localization, and incorrect class predictions, respectively. Compared to Source-Only and Vanilla Mean-Teacher, RT-SFOD successfully recovers missed objects, refines bounding box localization, and predicts the correct object categories.}
    \label{fig:qual_comparison}
    \vspace{-16pt}
\end{figure}

In Fig. \ref{fig:qual_comparison}, we provide a qualitative comparison on the Foggy Cityscapes validation set between \textit{Source-Only}, \textit{Vanilla Mean-Teacher (o2o)}, and our \textbf{RT-SFOD} under severe fog-induced domain shift. The Source-Only detector frequently misses objects and produces unreliable predictions, indicating poor robustness when deployed without target adaptation. While vanilla Mean-Teacher improves over the Source-Only baseline, it remains suboptimal, still exhibiting missed instances and imprecise bounding box localization. In contrast, RT-SFOD produces noticeably cleaner detections by improving both coverage and quality of the predictions. Although only the one-to-one (O2O) head is used during validation, the superior predictions of RT-SFOD stem from its improved training strategy. In particular, DHF enhances the quality and completeness of pseudo-labels during training by effectively leveraging complementary supervision from dual heads, leading to more reliable target adaptation. This improved pseudo-supervision enables the student model to learn stronger representations, resulting in better object recovery and tighter localization at inference time. Furthermore, MARD mitigates the domain-shift induced degradation of multi-scale feature discriminability by enforcing variance and covariance regularization on detection-aware foreground and background features across pyramid levels. Together, these components allow RT-SFOD to produce more accurate and complete detections.

\subsection{RT-SFOD is Robust to Partial Visibility}
\label{sec:partial_visibility}
\begin{figure}[t]
    \centering
    \includegraphics[width=\linewidth]{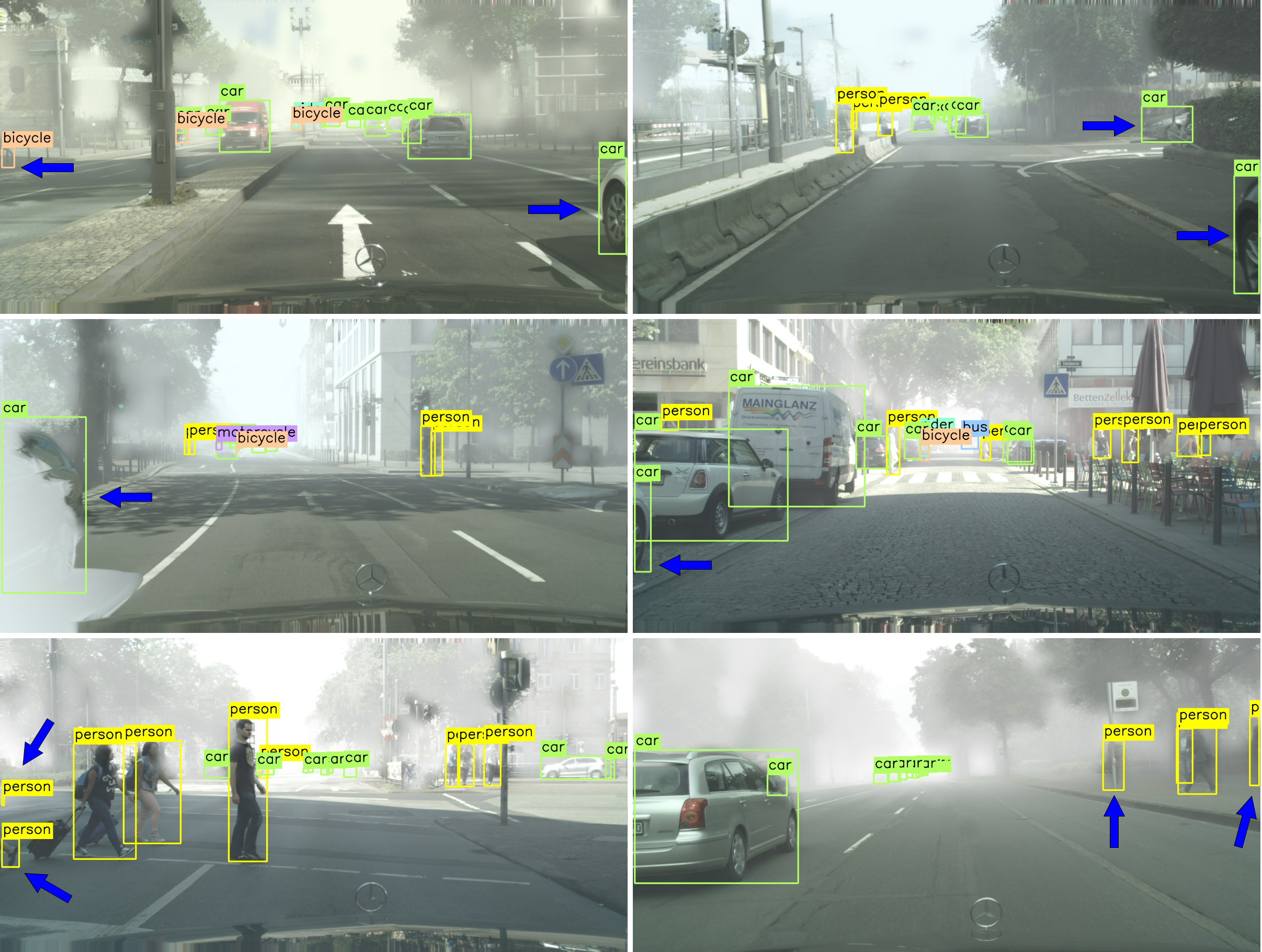}
    \caption{Robust detection of partially visible objects by \textbf{RT-SFOD} on the Foggy Cityscapes validation set. The model successfully detects heavily occluded and truncated objects under adverse visibility conditions. The \textcolor{blue}{blue} arrows highlight the partially visible objects.}
    \label{fig:partial_visibility}
    \vspace{-14pt}
\end{figure}

Fig. \ref{fig:partial_visibility} demonstrates the robustness of RT-SFOD in detecting partially visible and truncated objects in challenging foggy urban scenes. 
As highlighted in the examples (Fig. \ref{fig:partial_visibility}), objects that are only marginally visible due to occlusion, truncation at image boundaries, or severe fog are still reliably detected by the model. 
Such scenarios are particularly demanding because only limited visual cues are available, requiring strong feature generalization and contextual understanding. 
The consistent detection of these partially observable objects indicates that RT-SFOD learns discriminative representations that remain effective even when object appearance is incomplete. 
This capability is especially critical for real-time applications such as autonomous driving, where early and reliable detection of partially visible road users can significantly enhance safety and decision-making.

\section{Pseudo-label Quality Over Training Epochs}
\label{sec:pl_quality}
\begin{figure}[h]
    \centering
    \includegraphics[width=0.8\linewidth]{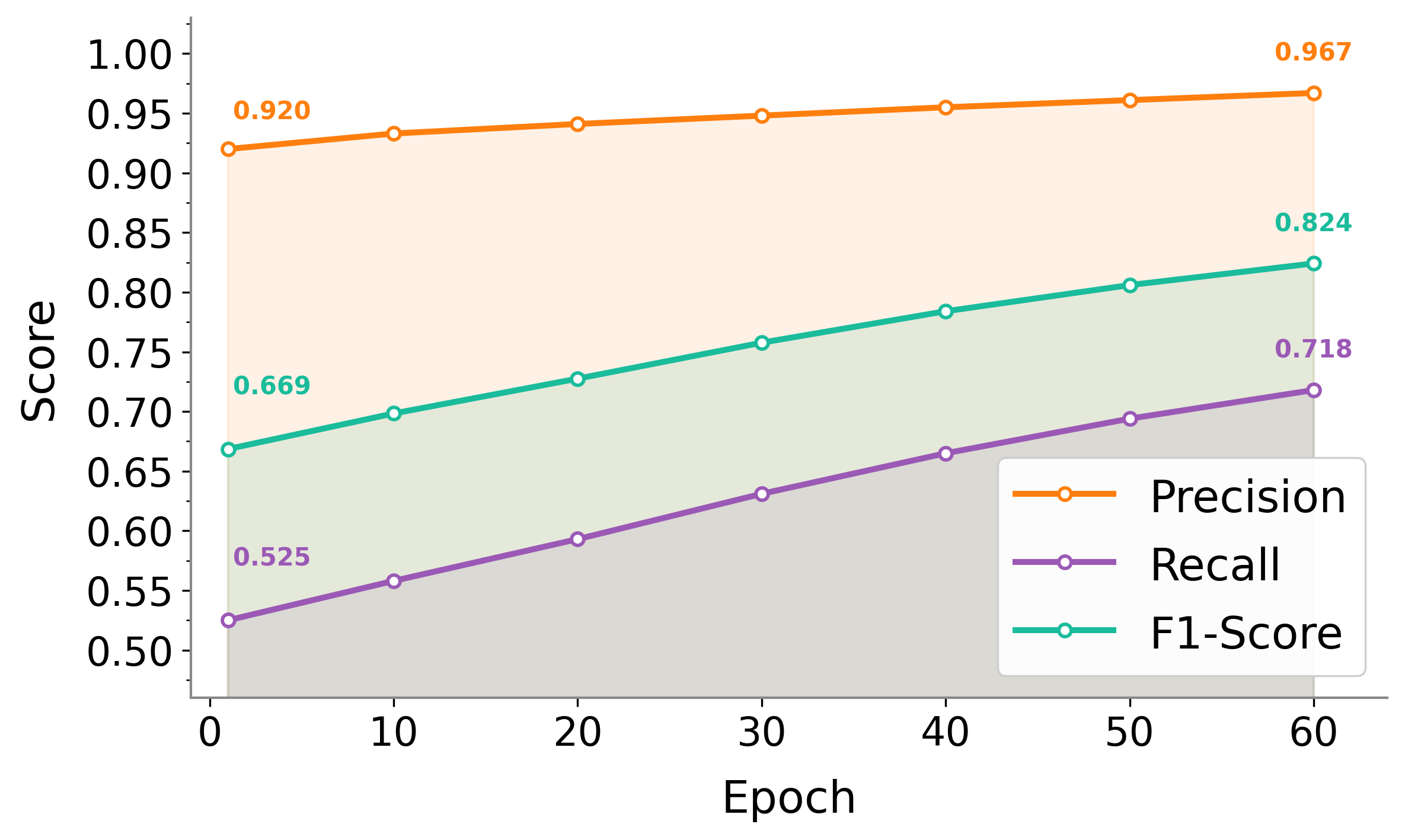}
    \caption{Evolution of pseudo-label quality generated by \textbf{DHF} over training epochs. Precision, recall, and F1-score consistently improve as adaptation progresses, indicating stable and progressively refined pseudo supervision.}
    \label{fig:prec_recall_supply}
    \vspace{-22pt}
\end{figure}
Fig. \ref{fig:prec_recall_supply} analyzes the evolution of pseudo-label quality produced by DHF throughout the adaptation process. 
We report precision, recall, and F1-score across training epochs to evaluate the reliability and completeness of the generated pseudo-labels. 
Precision starts at a high value and further improves steadily, demonstrating that DHF maintains strong noise control while adaptation proceeds. 
More importantly, recall increases consistently over epochs, indicating that the model progressively discovers additional valid target instances as its representation becomes more domain-aligned. 
The resulting F1-score exhibits a smooth and monotonic upward trend, reflecting a balanced improvement in both accuracy and coverage. 
This behavior confirms that the proposed framework is stable during training and exhibits a self-improving mechanism.

\section{Reproducibility}
\label{sec:reproducibility_supply}
In this section, we provide the hyperparameter details of RT-SFOD, dataset descriptions, and details of the augmentations used. 

\subsection{Hyperparameter analysis}
\label{subsec:hyperparams}
\begin{figure}[H]
    \centering
    \includegraphics[width=\linewidth]{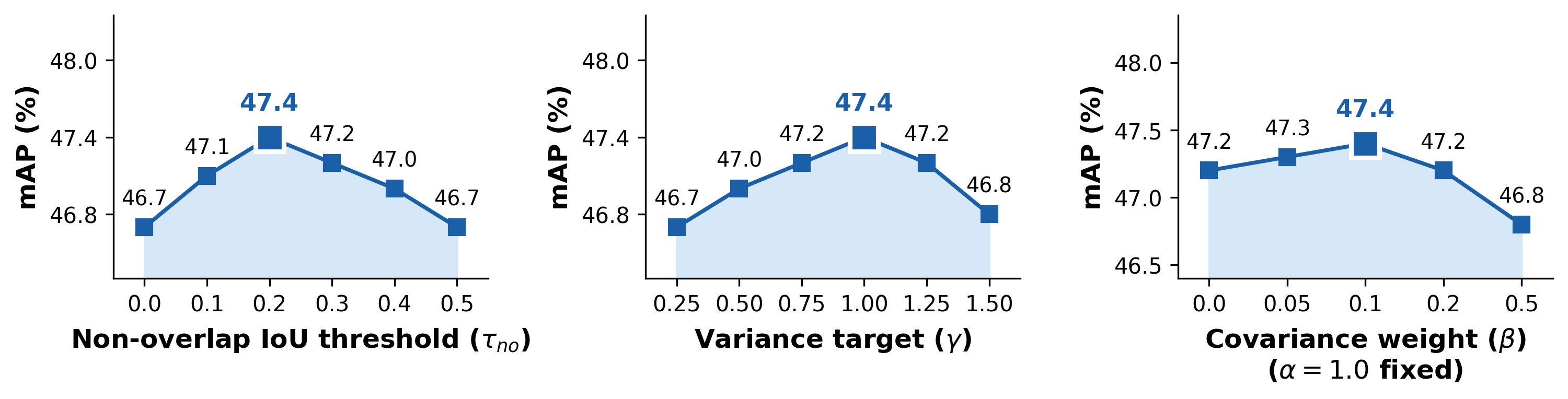}
    \vspace{-14pt}
    \caption{\textbf{Hyperparameter sensitivity} on C2F (RT-SFOD-M). mAP (\%) vs. (\textit{left}) the DHF non-overlap IoU threshold $\tau_{no}$, (\textit{center}) the MARD variance target $\gamma$, and (\textit{right}) the MARD covariance weight $\beta$ with $\alpha$ fixed at 1.0. Variation is $\leq$0.7 mAP in all cases.}
    \label{fig:supp_hpy}
\end{figure}

We provide the hyperparameter values used by our method in Table \ref{tab:hyperparams}. We analyze four key hyperparameters in the main paper (Fig. 4) and three additional ones ($\tau_{no}$, $\gamma$, and $\beta$) in Fig. \ref{fig:supp_hpy}. We observe that the remaining hyperparameters exhibit negligible sensitivity ($\leq$0.3 mAP variation) over their tested ranges.
As shown in Fig. \ref{fig:supp_hpy}, performance varies by at most 0.7 mAP across all tested values of $\tau_{no}$, $\gamma$, and $\beta$, confirming that the method does not depend on careful tuning for stable performance. 
Importantly, the \emph{same} set of hyperparameters achieved consistent gains across all four domain-shift benchmarks (Cityscapes $\rightarrow$ Foggy Cityscapes, Cityscapes $\rightarrow$ BDD100k, Sim10k $\rightarrow$ Cityscapes, and KITTI $\rightarrow$ Cityscapes), spanning diverse shift types including weather, large-scale scene diversity, synthetic-to-real rendering gaps, and camera shift.
This cross-benchmark stability indicates that RT-SFOD exhibits robustness and low sensitivity to its hyperparameter choices.

\begin{table}[t]
\centering
\caption{Complete hyperparameter settings for RT-SFOD. The same values are
used across all four domain-shift benchmarks. Sensitivity analysis plots are
provided for a subset of hyperparameters; remaining ones show negligible ($\leq$0.3 mAP)
variation over their tested ranges.}
\label{tab:hyperparams}
\setlength{\tabcolsep}{5pt}
\begin{tabular}{llccc}
\toprule
\textbf{Group} & \textbf{Hyperparameter} & \textbf{Symbol} & \textbf{Value} & \textbf{Sensitivity} \\
\midrule
\multirow{4}{*}{DHF}
  & O2O confidence threshold     & $\tau_{o2o}$  & 0.5   & Fig. 4 (main paper)       \\
  & O2M confidence threshold     & $\tau_{o2m}$  & 0.5   & Fig. 4 (main paper)      \\
  & Non-overlap IoU threshold    & $\tau_{no}$   & 0.2   & Fig. \ref{fig:supp_hpy}     \\
  & O2M NMS IoU threshold        & $\tau_{dup}$  & 0.7   & negligible   \\
\midrule
\multirow{11}{*}{MARD}
  & Base regularizer weight      & $\lambda_0$            & 0.05  & Fig. 4 (main paper)     \\
  & Regularizer weight clamp     & $\lambda_{max}$        & 0.2   & negligible \\
  & Variance target              & $\gamma$               & 1.0   & Fig. \ref{fig:supp_hpy}   \\
  & Variance term weight         & $\alpha$               & 1.0   & negligible \\
  & Covariance term weight       & $\beta$                & 0.1   & Fig. \ref{fig:supp_hpy}   \\
  & Warmup duration              & $T_{warmup}$           & 5 ep  & negligible \\
  & Confidence gate threshold    & $\bar{s}_{gate}$       & 0.5   & negligible \\
  & Top-$K_b$ boxes for sampling & $K_b$                  & 15    & negligible \\
  & Foreground points per box    & $K$                    & 8     & negligible \\
  & Background sample points     & $M$                    & 128   & negligible \\
  & Level assignment threshold   & $\eta$                 & 12.0  & negligible \\
\midrule
{EMA}
  & Teacher momentum             & $\mu$                  & 0.999 & Fig. 4 (main paper)     \\
\bottomrule
\end{tabular}
\end{table}

\subsection{Dataset Details}
\label{subsec:dataset_details_supply}
We conduct experiments on five open-access benchmarks that represent four types of domain shift. The \emph{Cityscapes} dataset \cite{cityscapes} contains 5,000 high-quality annotated urban street images gathered from multiple cities and seasonal settings. From this collection, 2,925 images are used for training and 500 for validation. The dataset covers eight object classes: person, rider, car, truck, bus, train, motorcycle, and bicycle.
\emph{Foggy Cityscapes} \cite{foggy_cityscapes} is derived from Cityscapes by adding simulated fog with three different density levels (0.005, 0.01, and 0.02), creating conditions that mimic reduced visibility. Following prior work \cite{falcon-sfod, simple-sfod, irg}, we only use the 0.02 density level images for target adaptation.
\emph{Sim10k} \cite{sim10k} consists of 10,000 synthetic urban images featuring cars, produced using the Grand Theft Auto game engine.
The \emph{KITTI} dataset \cite{kitti} includes 7,481 annotated real-world driving images captured for autonomous driving research.
Similar to existing works \cite{irg, pets, dru, simple-sfod, sf-yolo, falcon-sfod, beyond_boundaries}, for Sim10k $\to$ Cityscapes and KITTI $\to$ Cityscapes, we consider just the car class and report AP (on Car).
\emph{BDD100k} \cite{bdd100k} is a large-scale driving dataset with 100,000 images collected under diverse environmental and weather conditions.
\subsection{Details of Data Augmentation Strategies}
\label{subsec:details_data_aug_supply}
To facilitate robust self-training in our Source-Free Object Detection (SFOD) framework, we employ a Teacher-Student distillation mechanism with asymmetric data augmentations. This section details the specific transformations applied to the target domain images.

\textbf{Weak Augmentation ($\mathcal{A}_{weak}$)}
The teacher model receives a weakly augmented version of the target image to ensure the stability and reliability of the generated pseudo-labels. We avoid stochastic geometric distortions here to maintain spatial consistency. The operations include:
\begin{itemize}
    \item \textbf{Resize:} Images are resized such that the longest edge is $1024$ pixels, maintaining the original aspect ratio. 
    \item \textbf{Horizontal Flip:} Random horizontal flipping is applied with a probability of $p=0.5$.
\end{itemize}

\textbf{Strong Augmentation ($\mathcal{A}_{strong}$)}
The student model is tasked with predicting the teacher's pseudo-labels under heavy perturbation. This forces the model to rely on intrinsic object features rather than domain-specific noise or lighting conditions. The transformations include:

\begin{enumerate}
    \item \textbf{Geometric Transformations:} In addition to weak augmentations, we apply random affine transformations (scaling between $[0.9, 1.1]$ and translation up to $\pm 5\%$) and perspective distortions to simulate viewpoint changes.
    \item \textbf{Photometric Jitter:} We apply random Hue-Saturation-Value (HSV) adjustments, contrast/brightness shifts, Gamma corrections, and RGB channel shuffling to simulate varying sensor characteristics and lighting.
    \item \textbf{Noise and Blur:} To mimic low-quality or out-of-focus captures, we inject Gaussian noise, salt-and-pepper noise, and Gaussian blur.
\end{enumerate}

\begin{table}[h]
\centering
\caption{Data Augmentations used in RT-SFOD Training. \textit{Shared} refers to the ones used in both weak and strong augmentations.}
\label{tab:aug_params}
\begin{tabular}{@{}lll@{}}
\toprule
\textbf{Category} & \textbf{Augmentation} & \textbf{Range} \\ \midrule
\textit{Shared} & Resize & $1024 \times 1024$ (Long-edge) \\
 & Horizontal Flip & $p=0.5$ \\ \midrule
\textit{Geometric} & Scale/Translate & $p=0.5$, scale $\in [0.9, 1.1]$, trans. $\pm 5\%$ \\
 & Perspective & $p=0.3$, distortion $\pm 3\%$ \\ \midrule
\textit{Photometric} & HSV Jitter & $p=0.8$, H: $\pm 0.15$, S/V: $\pm 0.2$ \\
 & Contrast/Brightness & $p=0.6$, $\alpha \in [0.8, 1.2]$, $\beta \in [-20, 20]$ \\
 & Gamma Correction & $p=0.3$, $\gamma \in [0.7, 1.3]$ \\ 
 & Channel Shuffle & $p=0.2$, order $\in \{R, G, B\}$ permutations \\ \midrule
\textit{Degradation} & Gaussian Blur & $p=0.4$, kernel $\in \{3, 5, 7\}$, $\sigma \in [0.5, 2.0]$ \\
 & Noise (Gaussian) & $p=0.3$, $\sigma_{std} \in [5, 15]$ \\ 
 & Noise (Salt \& Pepper) & $p=0.3$, $prob \in [0.01, 0.05]$ \\ \midrule
\end{tabular}
\end{table}


\section{Failure-case Analysis}
\label{sec:failure_case_supply}
\begin{figure}[t]
    \centering
    \includegraphics[width=\linewidth]{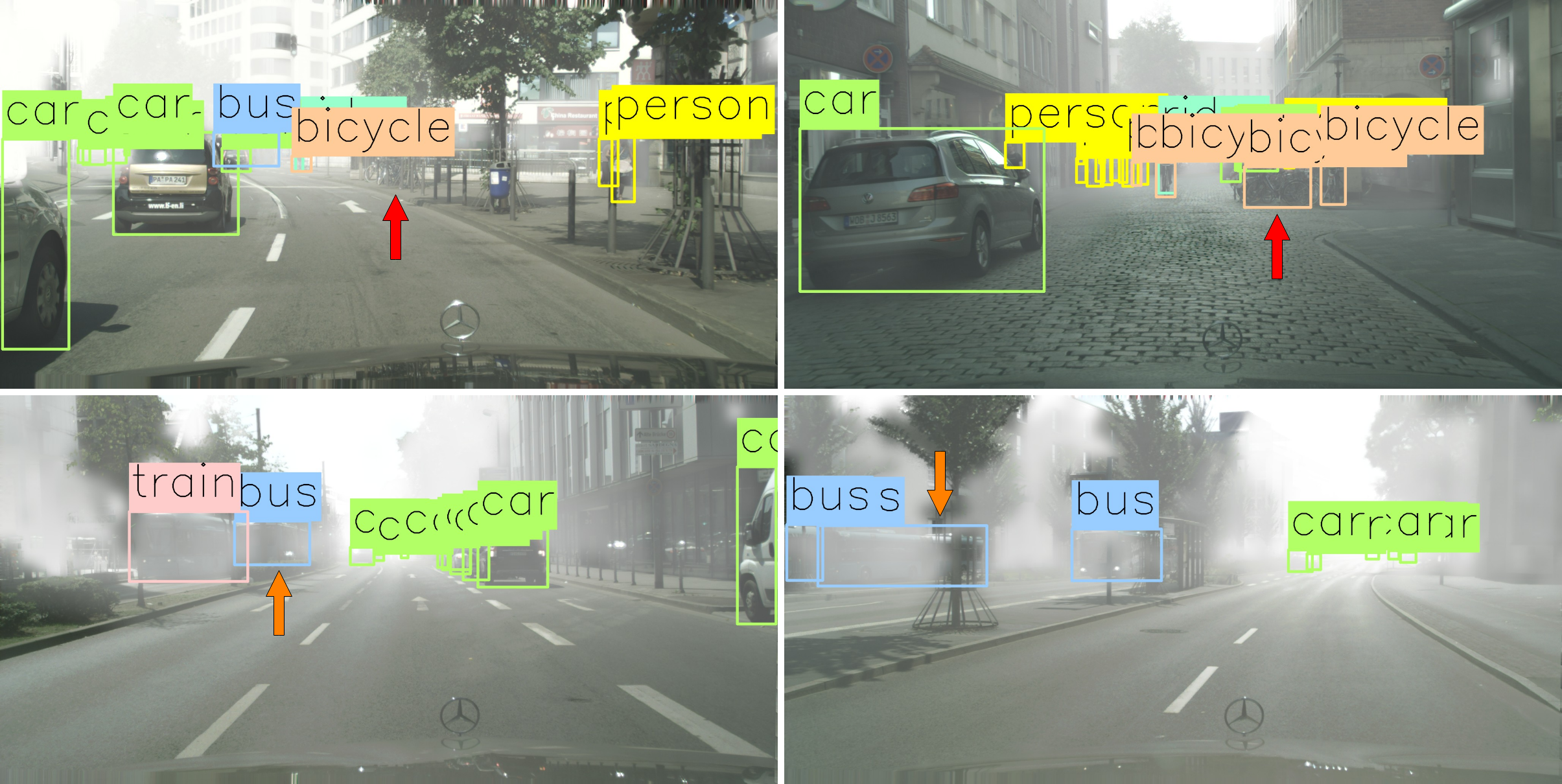}
    \caption{Failure cases of \textbf{RT-SFOD} on the Foggy Cityscapes validation set under extremely challenging conditions. Typical errors occur in crowded small-object scenarios and under severe fog and overexposure, where visibility is significantly degraded.}
    \label{fig:failure_cases}
\end{figure}
Fig. \ref{fig:failure_cases} illustrates challenging scenarios where RT-SFOD encounters limitations under extreme visual degradation. In the first row, densely packed and small objects, such as bicycles, create heavy occlusion and overlap. In such highly crowded regions, the detector may miss a few instances or produce a single bounding box covering multiple objects due to the limited spatial separability at that scale. The second row presents cases with extremely dense fog and strong sunlight exposure, where object boundaries are severely washed out, and contrast is drastically reduced, sometimes making objects barely distinguishable even to the human eye. Under these rare but severe conditions, RT-SFOD may produce suboptimal localization. Nevertheless, these examples highlight the inherent difficulty of the scenarios rather than the systematic weaknesses of the method. Overall, RT-SFOD remains robust across the vast majority of target-domain images and demonstrates strong resilience to domain shift, with failures primarily occurring in extreme edge cases involving severe visibility degradation and dense small-object clustering.

\section{Teacher EMA Update Variations}
\label{sec:ema_variation}
A known failure mode of fully-unlabeled teacher-student self-training is catastrophic collapse, where degraded teacher predictions are amplified through the pseudo-label feedback loop and propagate to the student. PETS \cite{pets} mitigates this via stronger temporal separation (periodic teacher–student exchange and an additional slower teacher), highlighting the importance of keeping the teacher as a stable target rather than a rapidly moving copy of the student. Simple-SFOD \cite{simple-sfod} shows that Adaptive Batch Normalization (AdaBN) provides a better starting point on the target domain and makes fixed pseudo-label training competitive with EMA-based methods. However, AdaBN does not eliminate pseudo-label noise or the non-stationarity introduced by augmentation and thresholding. 

\begin{figure}[t]
    \centering
    \includegraphics[width=0.8\linewidth]{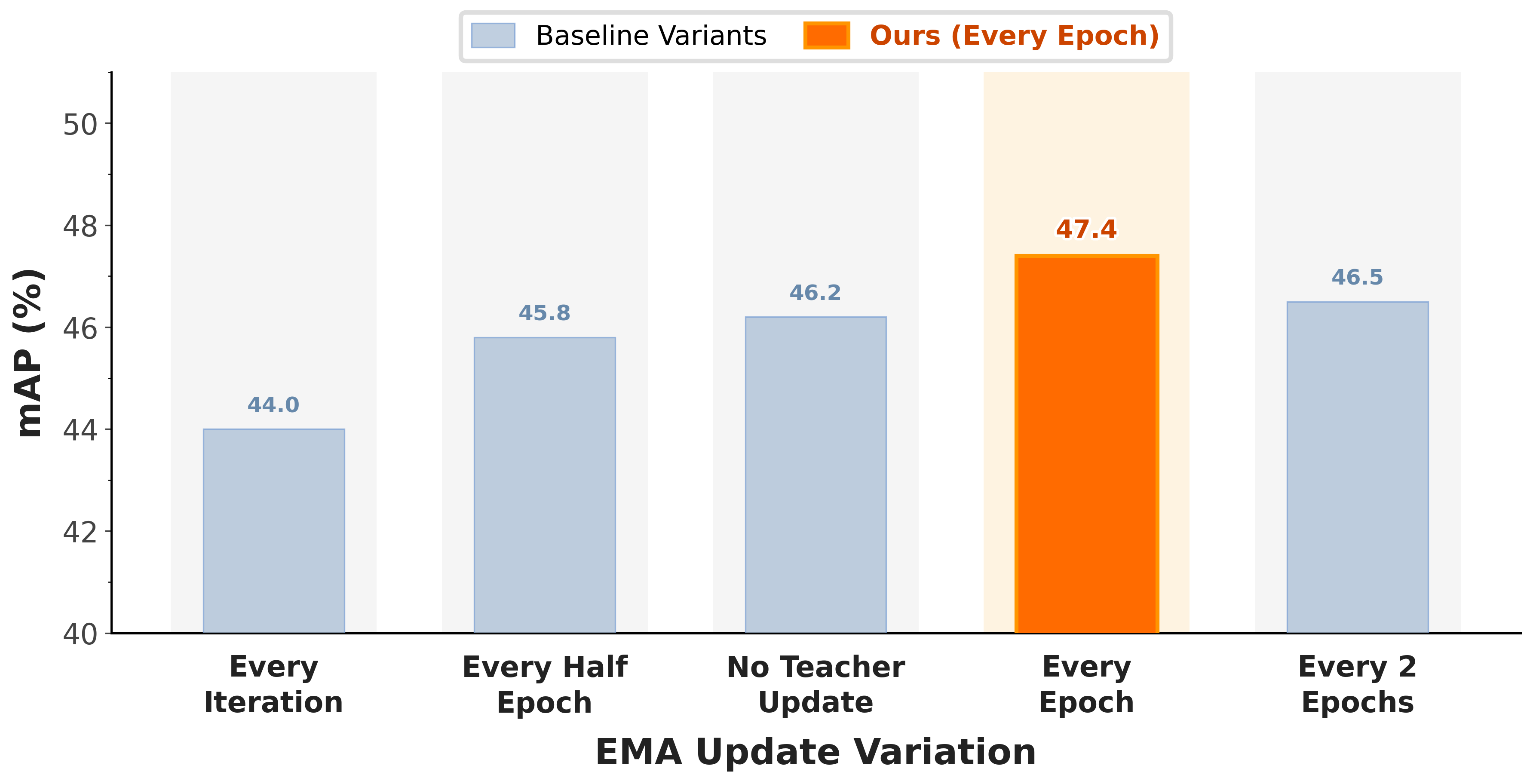}
    \caption{Impact of EMA teacher update frequency (momentum = 0.999) on target-domain performance (mAP\%) - C2F domain-shift. Updating the teacher once per epoch achieves the best performance (47.4\%), outperforming per-iteration updates, less frequent updates, and a frozen teacher (no update).}
    \label{fig:ema_variations}
    \vspace{-22pt}
\end{figure}

Figure \ref{fig:ema_variations} studies the effect of teacher update frequency under a fixed EMA momentum of 0.999 on our RT-SFOD-M. Updating the teacher at every iteration results in lower performance (44.0\%), indicating that overly frequent updates cause the teacher to closely track short-term student fluctuations, which may amplify noisy pseudo-labels. In contrast, a frozen teacher (no update) achieves 46.2\%, suggesting that a stable target is beneficial but limited by its inability to improve pseudo-label quality over time. Interestingly, updating every half epoch (45.8\%) underperforms a frozen teacher (46.2\%), indicating that intermediate update frequencies may introduce pseudo-label drift without sufficient temporal separation to stabilize training. Updating the teacher once per epoch achieves the best performance (47.4\%), outperforming both more frequent and less frequent updates. This schedule maintains sufficient temporal separation to avoid reinforcing transient student errors, while still allowing the teacher to progressively incorporate improved target-adapted representations. Updating every two epochs (46.5\%) slightly degrades performance, indicating that infrequent updates reduce the benefits of teacher adaptation. 

\section{Analysis of MARD Module}
\label{sec:mard_analysis}

\begin{figure}[t]
  \centering
  \subfloat[Domain-shift reduces feature effective rank]{\includegraphics[width=0.48\linewidth]{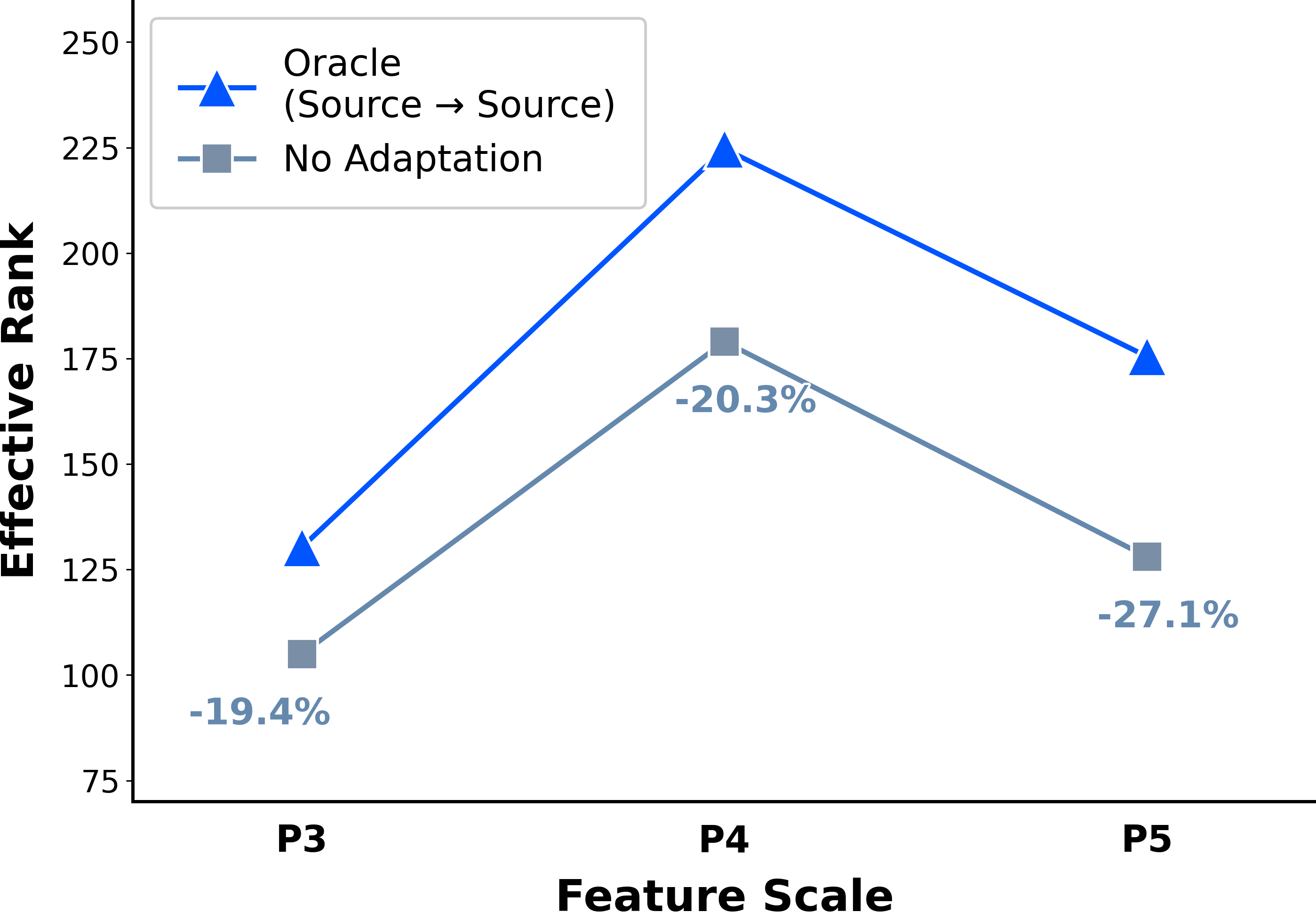}\label{fig:rank_collapse}}
  \hfill
  \subfloat[Rank recovery comparison]{\includegraphics[width=0.48\linewidth]{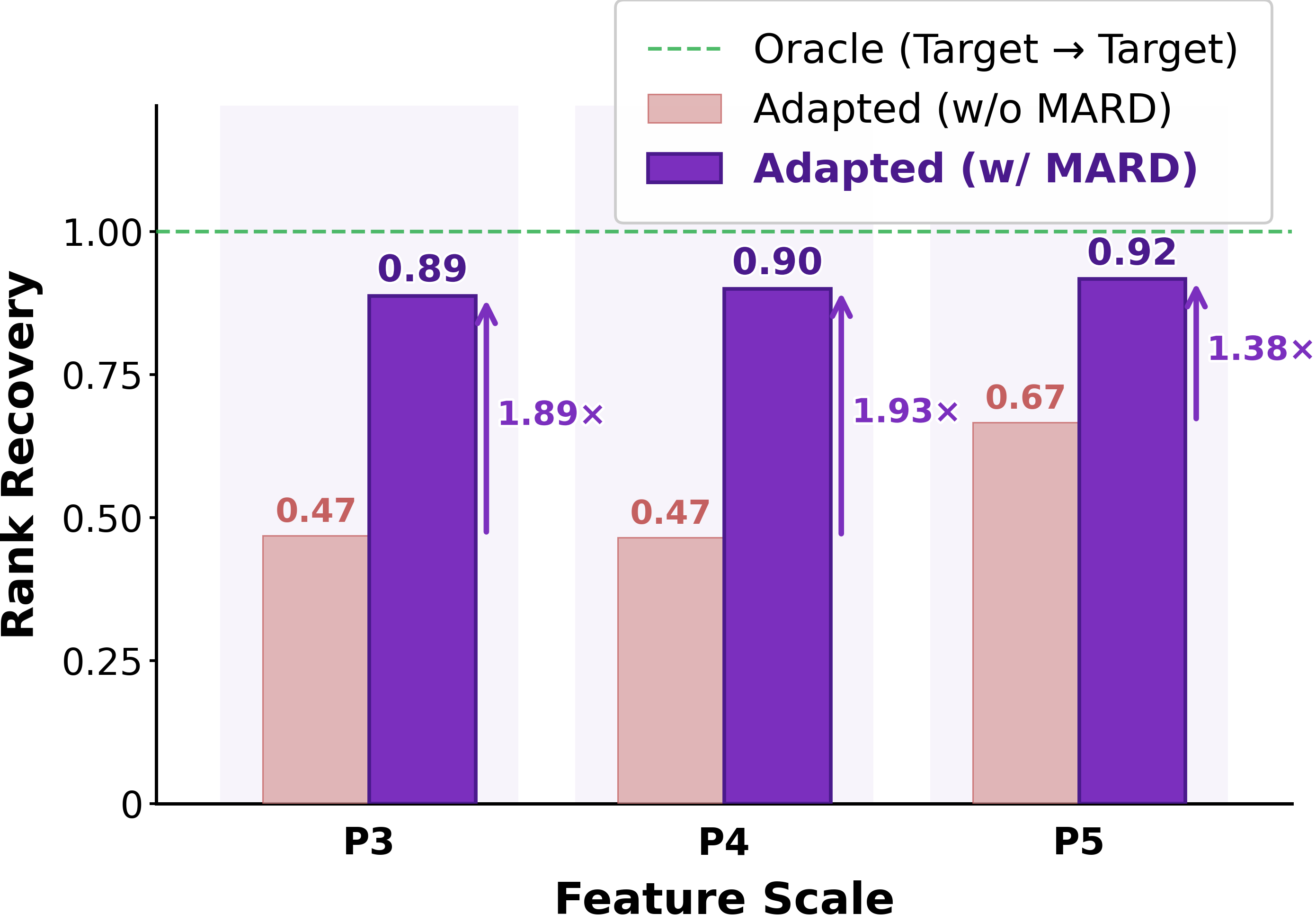} \label{fig:rank_recovery}}
  \caption{\textbf{Analysis of effective-rank degradation and recovery} on Cityscapes$\rightarrow$Foggy Cityscapes.
    (a) Domain shift reduces effective rank across feature scales.
    (b) A plot of Normalized rank recovery (Eq. \ref{eq:rank_recovery}) shows that MARD restores substantially more multi-scale feature diversity than vanilla mean-teacher self-training, closing the gap toward the target oracle.}
    \label{fig:rank_recovery_supp}
\end{figure}

In the MARD module, given $n$ channel-wise normalized tokens comprising $\tilde{Z}_\ell$, we define the covariance matrix for each level $\ell$ as: 
\begin{equation}
    \text{Cov}(\tilde{Z}_\ell) = \frac{1}{n-1} \sum_{i=1}^{n} (\tilde{z}_i - \bar{\tilde{z}})(\tilde{z}_i - \bar{\tilde{z}})^T, \quad \text{where} \quad \bar{\tilde{z}} = \frac{1}{n} \sum_{i=1}^{n} \tilde{z}_i.
\end{equation}
Here, $\text{Cov}(\tilde{Z}_\ell)$ represents the sample covariance matrix of the channel-wise normalized feature-vectors at the $\ell$-th level. The variable $n$ denotes the total number of feature-vectors, while $\tilde{z}_i$ corresponds to the $i$-th individual normalized feature-vector. The term $\bar{\tilde{z}}$ is the empirical mean of all $n$ feature-vectors within this level. The outer product $(\tilde{z}_i - \bar{\tilde{z}})(\tilde{z}_i - \bar{\tilde{z}})^T$ captures the pairwise feature correlations across the channels, and scaling by $\frac{1}{n-1}$ provides an unbiased estimate of the covariance matrix.

We define the channel wise variance as: 
\begin{equation}
    \text{Var}(Z_{\ell,:,j}) = \frac{1}{n-1} \sum_{i=1}^{n} (z_{i,j} - \bar{z}_j)^2, \quad \text{where} \quad \bar{z}_j = \frac{1}{n} \sum_{i=1}^{n} z_{i,j}.
\end{equation}
In this formulation, $\text{Var}(Z_\ell)_j$ denotes the sample variance of the $j$-th channel across all $n$ feature-vectors at level $\ell$. The term $z_{i,j}$ represents the activation of the $j$-th channel for the $i$-th feature-vector, and $\bar{z}_j$ is the empirical mean of that specific channel over all feature-vectors. 

In Fig. 2(b) (main paper), we quantify multi-scale feature discriminability using the \emph{effective rank} of the YOLOv10 PAN feature maps at levels $\ell\in\{P3,P4,P5\}$. We provide further supplemental analysis of MARD here. Similar to Fig.~2(b), \emph{Source Oracle}, refers to the source-trained model evaluated on the source validation set (Cityscapes$\rightarrow$Cityscapes); \emph{No Adaptation} refers to the source-trained model evaluated directly on the target validation set (Cityscapes$\rightarrow$Foggy Cityscapes); and the \emph{Target Oracle} refers to a fully supervised model trained and evaluated on the target domain (Foggy Cityscapes$\rightarrow$Foggy Cityscapes).

\noindent Fig. \ref{fig:rank_collapse} shows the phenomenon of a consistent drop in effective rank induced due to domain shift, across feature scales.

\noindent We define a normalized \emph{rank recovery} score at each scale $\ell$:
\begin{equation}
\mathrm{Recovery}_\ell
=
\frac{\mathrm{Rank}_\ell - \mathrm{NoAdapt}_\ell}{\mathrm{TargetOracle}_\ell - \mathrm{NoAdapt}_\ell},
\label{eq:rank_recovery}
\end{equation}
where $\mathrm{Rank}_\ell$ denotes the effective rank obtained by a given adaptation variant at scale $\ell$,
$\mathrm{NoAdapt}_\ell$ is the effective rank of the source model on the target domain (no adaptation),
and $\mathrm{TargetOracle}_\ell$ is the effective rank of a target-supervised oracle on the target domain.
This normalization expresses recovery as the fraction of the \emph{recoverable gap} from $\mathrm{NoAdapt}_\ell$ to the target oracle: 
$\mathrm{Recovery}_\ell=0$ corresponds to no improvement over $\mathrm{NoAdapt}_\ell$, and $\mathrm{Recovery}_\ell=1$ corresponds to fully closing the gap to the target oracle. Fig. \ref{fig:rank_recovery} plots $\mathrm{Recovery}_\ell$ for vanilla mean-teacher adaptation and for our proposed MARD. Consistent with Fig. 2(b), vanilla self-training recovers only a limited portion of rank, whereas MARD recovers substantially more rank at all pyramid levels, demonstrating that it effectively counters the domain-shift-induced feature collapse, hence preserving multi-scale feature discriminability.

\section{TensorRT FP16 Efficiency Evaluation}
\label{sec:trt_fp16}

As discussed in the main paper (Sec. 4), prior efficiency-focused object detection works \cite{rtdetr, sf-yolo, yolov10} frequently report latency using TensorRT FP16, while reporting accuracy in FP32 PyTorch. To facilitate reproducibility and provide a reference under this commonly reported deployment setting, we additionally report TensorRT FP16 latency for RT-SFOD in Table \ref{tab:trt_fp16_latency}. These results are provided for completeness and comparability only. All primary comparisons in the main paper remain based on FP32 PyTorch evaluation.

\noindent \textbf{Experimental Protocol.}
We measure latency using TensorRT FP16 on a Tesla T4 GPU. Models are exported with an input resolution of $1 \times 3 \times 512 \times 1024$ and batch size 1. Timing is performed using CUDA events after a warmup phase of 100 iterations, followed by 500 measured iterations. We report forward-pass execution time only, excluding image decoding, preprocessing, host-device transfer, and post-processing. The statistics p50, p90, and p99 denote the 50th, 90th, and 99th percentile latency values across measured iterations, respectively.

\begin{table}[t]
\centering
\caption{\textbf{TensorRT FP16 forward-pass latency} on a Tesla T4 GPU, batch size 1, input resolution $512 \times 1024$.}
\small
\begin{tabular}{lccccc}
\toprule
\textbf{Model} & \textbf{Mean (ms)} & \textbf{p50} & \textbf{p90} & \textbf{p99} & \textbf{FPS} \\
\midrule
RT-SFOD-S & 3.611 & 3.633 & 3.808 & 6.559 & 276.93 \\
RT-SFOD-M & 7.795 & 7.769 & 7.926 & 8.286 & 128.29 \\
RT-SFOD-L & 12.253 & 12.187 & 12.558 & 12.745 & 81.61 \\
\bottomrule
\end{tabular}
\label{tab:trt_fp16_latency}
\end{table}

These numbers reflect the efficiency of RT-SFOD, which reduces to a standard YOLOv10 detector during inference. The reported latency is not directly comparable to the FP32 PyTorch efficiency values in the main paper due to differences in precision, runtime framework, hardware, and measurement protocol.


\section{Training Overhead of Proposed Modules}
\label{sec:training_overhead}
\vspace{-22pt}
\begin{table}[H]
\centering
\caption{Training overhead of proposed modules relative to Vanilla 
Mean-Teacher (RT-SFOD-M, Cityscapes~$\to$~Foggy Cityscapes, 
single NVIDIA RTX A6000 GPU, batch size 16).}
\label{tab:overhead}
\resizebox{\columnwidth}{!}{%
\begin{tabular}{lcccc}
\toprule
\textbf{Configuration} 
    & \textbf{Peak Mem. (MB)} & \textbf{$\Delta$ Mem.} 
    & \textbf{Time/Epoch (s)} & \textbf{$\Delta$ Time} \\
\midrule
Vanilla MT (baseline)       & 21,120 & ---     & 139.0 & --- \\
\quad + DHF                 & 21,400 & +280    & 145.4 & +6.4\,s\ (+4.6\%) \\
\quad + MARD                & 22,360 & +1,240  & 156.0 & +17.0\,s\ (+12.2\%) \\
\quad + DHF + MARD (\textbf{RT-SFOD}) & 22,590 & +1,470 & 162.7 & +23.7\,s\ (+17.1\%) \\
\bottomrule
\end{tabular}%
}
\end{table}

Table. \ref{tab:overhead} reports the peak GPU memory and time per epoch for 
each configuration, measured on a single NVIDIA RTX A6000 GPU with batch 
size 16 on the C2F benchmark. Both DHF and MARD are \textit{training-time 
only} modules that do not affect the deployed model, so this overhead has no 
bearing on inference FPS or latency reported in the main paper.

DHF introduces negligible cost: +280\,MB (+1.3\%) in peak memory and 
+6.4\,s (+4.6\%) per epoch, with no architectural changes to the student or 
teacher networks. The slightly higher overhead of MARD (+1,240\,MB, 
+17.0\,s over baseline) stems from computing per-level variance and covariance 
statistics over foreground and background feature vectors sampled from the 
multi-scale PAN outputs. The full RT-SFOD configuration (DHF + MARD) incurs a combined 
overhead of +1,470\,MB and +23.7\,s (+17.1\%) per epoch over Vanilla 
Mean-Teacher, a modest training-time cost in exchange for consistent 
1.4 to 3.5\% mAP gains across four benchmarks. The overhead is consistent 
across all domain-shift benchmarks.

\section{Additional Analyses and Clarifications}
\label{sec:additional}
This section consolidates additional experiments and clarifications that further support the claims made in the main paper.

\subsection{Generality Across Dual-Head Detectors}
\label{sec:supp_generality}
RT-SFOD targets NMS-free \emph{dual-head} detectors in general, not YOLOv10 specifically (Contribution~1, main paper). DHF requires only a precise one-to-one (O2O) head and a dense, high-recall one-to-many (O2M) head---an increasingly common design that combines strong supervision with NMS-free efficient inference. Representative examples include H-DETR (CVPR'23), MS-DETR (CVPR'24), RT-DETRv3 (WACV'25), Mr.\,DETR (CVPR'25), and YOLOv26. To verify that our findings are not tied to a single architecture, we apply RT-SFOD with the \emph{same} hyperparameters to three additional dual-head detectors on the Cityscapes~$\to$~Foggy Cityscapes (C2F) benchmark.

\vspace{-18pt}
\begin{table}[h]
\centering
\caption{Additional C2F results vs.\ the MT(O2O) baseline on YOLOv26S/M/L, MS-DETR (CVPR'24), and Mr.\,DETR (CVPR'25), using the same hyperparameters as YOLOv10.}
\label{tab:supp_generality}
\small
\begin{tabular}{lcc}
\toprule
\textbf{Detector} & \textbf{MT (O2O)} & \textbf{RT-SFOD} \\
\midrule
YOLOv26S  & 42.1 & 44.6 (+2.5) \\
YOLOv26M  & 46.0 & 48.8 (+2.8) \\
YOLOv26L  & 50.6 & 52.7 (+2.1) \\
MS-DETR   & 42.3 & 44.3 (+2.0) \\
Mr.\,DETR & 43.8 & 46.1 (+2.3) \\
\bottomrule
\end{tabular}
\end{table}

As shown in Table~\ref{tab:supp_generality}, RT-SFOD improves the MT~(O2O) baseline on YOLOv26S/M/L by +2.5/+2.8/+2.1 mAP, on MS-DETR by +2.0, and on Mr.\,DETR by +2.3, consistent with the gains observed on YOLOv10. This indicates that the precision--recall fusion exploited by DHF and the feature-diversification effect of MARD are properties of the dual-head paradigm rather than of one specific detector. For single-head detectors (e.g., YOLOv5/v11), DHF has no second candidate set and gracefully reduces to confidence-threshold pseudo-labeling as in SF-YOLO; MARD remains applicable whenever pseudo-labels and multi-scale features are available.

\subsection{Robustness to Cold-Start and Noisy MARD Sampling}
\label{sec:supp_robustness}
A natural concern with self-training under large domain gaps is error propagation: if the teacher's initial predictions are unreliable, the DHF anchors and the regions MARD samples could be corrupted from the outset.

\noindent \textbf{Cold start.} AdaBN warm-starts the teacher before self-training begins, and Fig.~\ref{fig:prec_recall_supply} shows that DHF pseudo-label precision is already $0.920$ at epoch~0, so the O2O anchors are reliable before adaptation. To probe a substantially larger initial gap than our driving benchmarks, we evaluate RT-SFOD-M on the extreme PASCAL~VOC~$\to$~Clipart shift, where it improves the MT~(O2O) baseline from $37.8$ to $40.5$ mAP.

\noindent \textbf{Noisy MARD sampling.} MARD mitigates noisy supervision through warm-starting, confidence-gating, and periodic application. To stress-test its sensitivity to inaccurate sampling regions directly, we corrupt the initial pseudo boxes on C2F (RT-SFOD-M) by jittering their centers and sizes for $20\%$/$40\%$/$60\%$ of boxes. mAP remains stable, changing only from $47.4$ to $47.2$/$47.0$/$46.9$, respectively. Together, these results show that RT-SFOD is robust both to severe cold-start shifts and to inaccurate early MARD sampling.


\subsection{Hyperparameter Selection and Transferability}
\label{sec:supp_hyper}
\noindent \textbf{MARD sampling counts.} The foreground/background sampling counts $K$/$M$ control how many feature vectors are drawn per pseudo box; the aim is to cover enough object and background regions without noisy over-sampling or excessive computational overhead. We swept $K\in\{2,4,8,16,32\}$ and $M\in\{16,32,64,128,256\}$ and found $K{=}8$, $M{=}128$ to give the best mAP, with negligible sensitivity ($\leq$0.3 mAP) over the tested ranges (Table~\ref{tab:hyperparams}).

\noindent \textbf{Transfer across scales and architectures.} We tuned hyperparameters once on YOLOv10-M and reused the \emph{same} values across all model scales and the additional dual-head detectors in Sec.~\ref{sec:supp_generality}, where they yield consistent gains. For example, varying $\tau_{o2o}$ over $\{0.3,0.4,0.5,0.6,0.7\}$ on YOLOv26S changes mAP by $\leq$1\%. We attribute this transferability to the dependence of our thresholds on \emph{relative} statistics rather than architecture-specific factors: confidence is internally calibrated (a threshold of 0.5 still selects reliable predictions in each detector), IoU is purely geometric, and domain shift affects multi-scale features similarly across detectors. As a result, a single shared configuration continues to work well across models and scales.

\subsection{Limitations}
\label{sec:supp_limitations}
While RT-SFOD advances the accuracy--speed--size trade-off for source-free detection, it has a few limitations. \textbf{(i) Architecture scope.} DHF is designed for NMS-free dual-head detectors; on single-head detectors it has no second candidate set and degenerates to simple confidence-threshold pseudo-labeling (as in SF-YOLO), so the pseudo-label fusion gains reported here are specific to the dual-head setting. MARD, however, remains applicable whenever pseudo-labels and multi-scale features are available. \textbf{(ii) Capacity on camera-shift.} On camera-shift benchmarks such as K2C, performance benefits from additional model capacity, as discussed in Sec.~\ref{sec:sota}. \textbf{(iii) Extreme visual degradation.} As shown in Fig.~\ref{fig:failure_cases}, residual failures occur under densely packed small objects and severe fog or overexposure, where visual cues are minimal. These cases reflect the inherent difficulty of the inputs rather than systematic weaknesses of the method, and RT-SFOD remains robust across the vast majority of target-domain images.

\end{document}